\documentclass{article}

\PassOptionsToPackage{numbers, compress}{natbib}

\usepackage{amsmath}
\usepackage{graphicx}
\usepackage{wrapfig}
\usepackage{multirow}
\usepackage{amsmath}
\usepackage{booktabs}
\usepackage{array}
\usepackage[dvipsnames]{xcolor}
\usepackage[preprint]{neurips_2026}

\usepackage[utf8]{inputenc} 
\usepackage[T1]{fontenc}    
\usepackage{hyperref}       
\usepackage{url}            
\usepackage{booktabs}       
\usepackage{amsfonts}       
\usepackage{nicefrac}       
\usepackage{xcolor}         
\usepackage{comment}

\usepackage[most]{tcolorbox}
\usepackage{listings}
\usepackage{xcolor}

\usepackage[most]{tcolorbox}

\newtcolorbox{finding}[1][]{
  colback=blue!3,
  colframe=blue!50!black,
  fonttitle=\bfseries,
  coltitle=white,
  colbacktitle=blue!50!black,
  title=#1,
  sharp corners=south,
  boxrule=0.6pt,
  left=6pt, right=6pt, top=4pt, bottom=4pt,
  fontupper=\itshape,
}

\newtcolorbox{examplebox}[1][]{
  enhanced,
  breakable,
  colback=gray!5,
  colframe=gray!55,
  boxrule=0.5pt,
  arc=1.5mm,
  left=2mm,
  right=2mm,
  top=1mm,
  bottom=1mm,
  fonttitle=\bfseries,
  title=#1
}

\newtcblisting{promptbox}[1][]{
  enhanced,
  breakable,
  listing only,
  colback=gray!3,
  colframe=gray!55,
  boxrule=0.5pt,
  arc=1.5mm,
  left=2mm,
  right=2mm,
  top=1mm,
  bottom=1mm,
  fonttitle=\bfseries,
  title=#1,
  listing options={
    basicstyle=\ttfamily\small,
    breaklines=true,
    columns=fullflexible,
    keepspaces=true
  }
}

\title{Overcoming State Inertia in Full-Duplex Spoken Language Models via Activation Steering}

\author{%
\begin{tabular}{c}
Cheng-Kuang Chang\thanks{Co-first authors.} \hspace{1.2em} 
Kai-Wei Chang\footnotemark[1] \hspace{1.2em}
Alexander H. Liu \hspace{1.2em}
James Glass \\
[0.5ex]\\
{\normalfont
MIT CSAIL
}
\\
\\
{\normalfont
\{chang168, kwchang\}@mit.edu
}
\end{tabular}
}

\begin{document}

\maketitle

\begin{abstract}
Full-duplex spoken language models (FD-SLMs) enable seamless speech interaction by allowing models to listen and speak simultaneously, yet the internal mechanism by which they coordinate listening and speaking remains underexplored. We analyze the predictive behavior encoded in FD-SLM hidden representations and find that they exhibit stream-specific predictive patterns: during listening, they preferentially predict the incoming user stream, whereas during speaking, they preferentially predict the model output stream. Building on this observation, we show that FD-SLMs dynamically modulate their internal predictive focus between two states: a \emph{generative state} aligned with model output generation and a \emph{perceptive state} aligned with incoming user input. However, this modulation can lag behind abrupt changes in conversational context. During user interruptions, the model remains transiently biased toward the generative state before transitioning into the perceptive state, causing it to miss the beginning of the incoming input. We term this delayed internal transition \textit{state inertia}. To quantify its downstream impact, we introduce the \emph{Zero-Buffer Benchmark (ZBB)}, a diagnostic benchmark for evaluating immediate interruption comprehension when user speech begins abruptly. We evaluate this setting using response correctness and initial-word occurrence rate (IWOR). Finally, we mitigate state inertia through activation steering with a \emph{perception vector}, a training-free intervention with little additional computational overhead. Across multiple state-of-the-art FD-SLMs, activation steering substantially improves interruption handling; for example, on PersonaPlex, it improves correctness from 28\% to 45\% and IWOR from 40\% to 72\% without any fine-tuning.
\end{abstract}

\section{Introduction}
Achieving human-level conversational fluency has long been a central goal in spoken dialogue systems~\cite{aroralandscape, glass1999challenges, ji2024wavchat}. Recently, \emph{full-duplex spoken language models (FD-SLMs)} have attracted increasing attention for their ability to listen and speak simultaneously, moving beyond the rigid turn-by-turn interaction of conventional half-duplex spoken language models (HD-SLMs)~\cite{chang2025game, chang2026tico, cui2025recent, veluri2024beyond, ji2024wavchat, lin2025full_v1, ding2025kimi, wu2025step, xu2025qwen3}. 
In practice, FD-SLMs often operate with a dual-channel structure~\cite{roy2026personaplex, defossez2024moshi, raon_speech_report, aroralandscape}, jointly processing a user stream containing incoming user speech and a model stream representing the model's own speech. This design enables timing-sensitive conversational behaviors such as backchanneling, smooth interruption handling, fluid turn-taking, and synchronized interaction~\cite{chang2025game, lin2025full_v1, lin2025full_v2, cui2025turn_guide}.

Despite these capabilities, the internal mechanism by which FD-SLMs coordinate listening and speaking remains underexplored. 
Inspired by \emph{logit lens}~\cite{nostalgebraist2020logitlens, belrose2023eliciting, rai2024practical}, we analyze the predictive behavior encoded in FD-SLM hidden representations. 
Our analysis reveals ``stream-specific'' predictive patterns: \emph{during listening, hidden representations preferentially predict the incoming user stream, whereas during speaking, they preferentially predict the model output stream}. 
We further find that \emph{FD-SLMs coordinate the listening and speaking behavior by dynamically modulating two states: the \textbf{``generative state''} and the \textbf{``perceptive state''}}.
However, this modulation is not always successful on demand. In particular, we find that when a user abruptly interrupts the model while it is speaking, the model remains transiently biased toward the generative state and fails to transition promptly into the perceptive state. We refer to this phenomenon as \textbf{``state inertia''}.

State inertia causes the model to miss the user input when an interruption occurs. This loss of information degrades the quality of the model's response.
Interestingly, ``state inertia'' resembles speech-induced suppression in human auditory processing, where speech production can suppress activity in the auditory cortex and increase auditory response latency~\cite{NUMMINEN1999119, houde2002modulation}.

To quantify the effect of state inertia, we introduce the \textbf{Zero-Buffer Benchmark (ZBB)}, a diagnostic benchmark for measuring whether FD-SLMs can immediately understand user input after interruption. Unlike existing benchmarks that evaluate overall dialogue quality~\cite{lin2025full_v1, peng2025fdbench, zhang2025mtrduplexbench, wang2026fullduplex_humdial_challenge}, ZBB places the critical semantic keyword as the first word of the interrupting utterance, with no leading filler or acoustic buffer~\cite{clark2002filler_word_uh, duvall2014exploring_filler_word}. This design directly tests whether the model perceives the earliest semantic information after interruption, precisely when state inertia is most likely to affect perception. We evaluate model performance using response correctness and Initial Word Occurrence Rate (IWOR), which measures whether the model recognizes the beginning of the interruption. Across multiple FD-SLMs, interruption substantially degrades both metrics, showing that state inertia has measurable behavioral consequences.

Finally, we mitigate state inertia using a training-free \emph{activation steering} method \cite{turner2023steering, zou2023representation, rimsky2024steering}. We construct a \emph{perception vector} from the difference between hidden representations in the generative state and the perceptive state, and apply it at the onset of interruption to steer the model toward the perceptive state. This steering requires no fine-tuning and adds only a lightweight inference-time hidden-state update. Empirically, steering with the perception vector consistently improves interruption handling across multiple FD-SLMs; for example, on PersonaPlex~\cite{roy2026personaplex}, it improves correctness from 28\% to 45\% and IWOR from 40\% to 72\%.

In summary, our main contributions are as follows:
\begin{itemize}
\item \textbf{Internal state analysis and state inertia:} We show that FD-SLM hidden representations exhibit stream-specific predictive behavior and dynamically modulate between generative and perceptive states. Building on this analysis, we identify \textit{state inertia}, a delayed internal transition that reduces the model's ability to process abrupt user interruptions.

\item \textbf{Zero-Buffer Benchmark (ZBB):} We introduce ZBB, a diagnostic benchmark for evaluating immediate interruption comprehension when user speech begins abruptly, together with correctness and Initial Word Occurrence Rate (IWOR).

\item \textbf{Training-free mitigation via activation steering:} We introduce a training-free activation steering method based on a perception vector, which mitigates state inertia and substantially improves interruption handling across multiple FD-SLMs.
\end{itemize}

\begin{figure}[htbp]
    \centering
    \includegraphics[width=\textwidth]{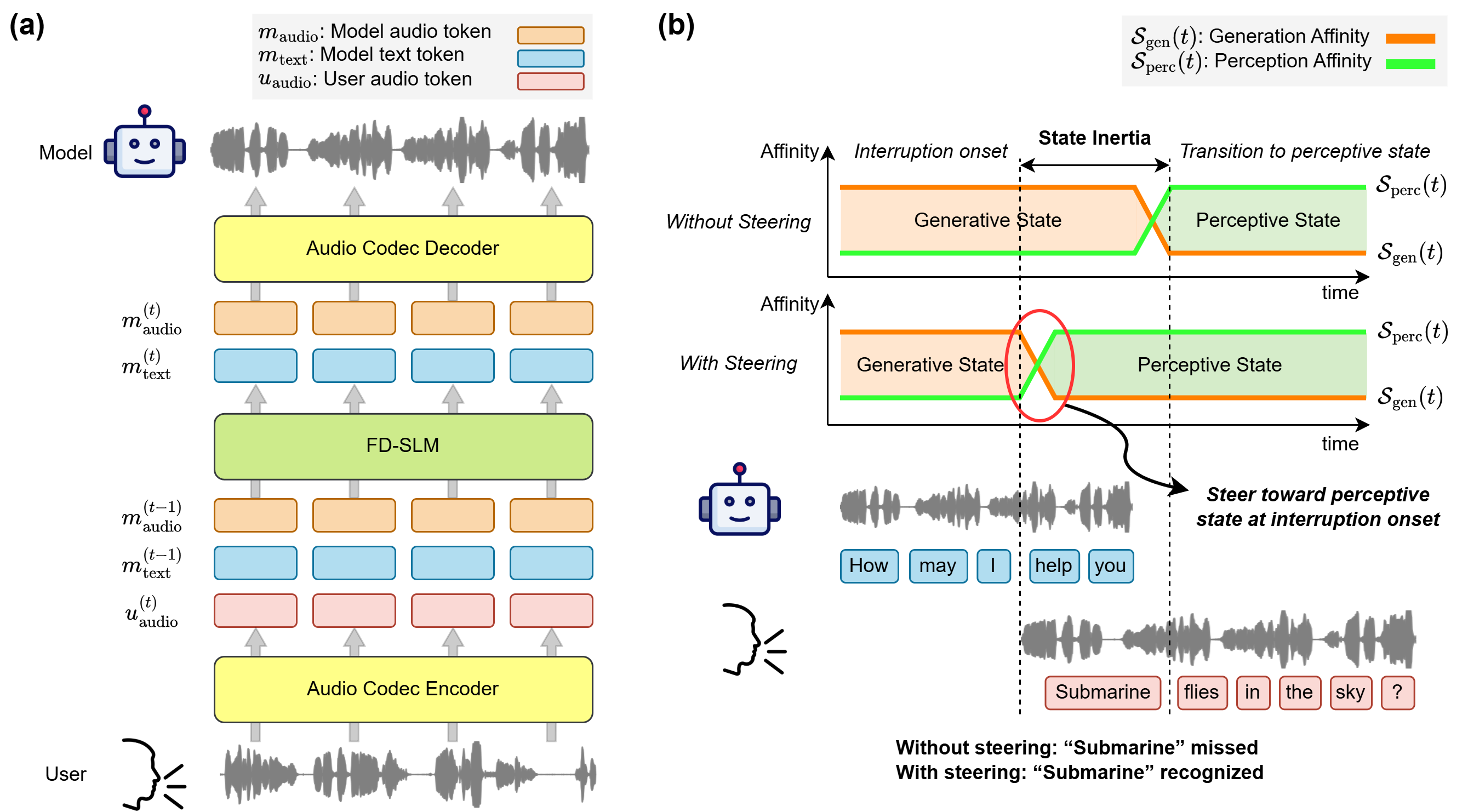}
    \caption{
    \textbf{Overview of state inertia and activation steering.}
    (a) FD-SLMs process concurrent user and model streams, conditioning on incoming user audio and previous model output tokens to generate text and audio tokens.
    (b) FD-SLMs coordinate speaking and listening by modulating between generative and perceptive states, tracked by generation and perception affinity. During abrupt interruptions, the model can remain biased toward the generative state before transitioning to the perceptive state, causing early user input to be missed. Injecting a perception vector at interruption onset accelerates this transition and improves interruption handling.
    }
    \label{fig:main}
\end{figure}

\section{Related Work}

\paragraph{Full-Duplex Spoken Language Models.}
Many existing spoken language models follow a half-duplex interaction pattern, processing input and output speech sequentially and relying on explicit turn-taking boundaries between listening and speaking~\cite{fang2025llamaomni, xie2024mini, zeng2024glm}. This rigid interaction pattern can make conversations feel unnatural, especially in scenarios involving interruptions, backchannels, or overlapping speech~\cite{skantze2021turn_taking_half_duplex}. In contrast, full-duplex spoken language models (FD-SLMs) support real-time bidirectional speech interaction, allowing the model to continuously perceive user audio while generating speech responses \cite{aroralandscape, veluri2024beyond, zhang2024full_duplex_def}. This capability enables more natural conversational behaviors, including backchanneling, interruption handling, and overlapping speech~\cite{lin2025full_v1}. Motivated by these advantages, recent work has developed several full-duplex systems, including open-source models such as Moshi~\cite{defossez2024moshi}, PersonaPlex~\cite{roy2026personaplex}, and Raon-SpeechChat \cite{raon_speech_report}. While these systems demonstrate the promise of full-duplex interaction, the internal mechanisms by which they coordinate simultaneous listening and speaking remain underexplored.

\paragraph{FD-SLMs Benchmarks.}
Existing benchmarks for FD-SLMs~\cite{ lin2025full_v2, lin2026full_v3, peng2025fdbench, chang2025game} primarily assess macroscopic conversational properties. These include turn-taking dynamics, such as properly taking or yielding the floor; end-to-end response latency; overall instruction following; and full-duplex-specific behaviors such as backchanneling. However, these benchmarks largely overlook a critical fine-grained capability: whether the model accurately recognizes user input immediately following an abrupt interruption. This distinction is important because a model may eventually recover and produce a plausible response while still missing information at the beginning of the interrupting utterance. In this work, we assess this moment-level listening ability, which we discuss in Section~\ref{sec:zbb_intro}.

\paragraph{Activation Steering.}
Activation steering modifies model behavior at inference time by injecting steering vectors into hidden states, often using mean-activation differences between contrasting concepts or behaviors~\cite{zou2023representation, turner2023steering, rimsky2024steering}. Prior work has used steering to control text-generation behavior, such as instruction following, persona modification, vulnerability analysis, and representation probing~\cite{stolfoimproving, chen2025persona, wang2024trojan_activation_attack, alain2017understanding}. We instead apply activation steering to FD-SLMs, using it to steer hidden representations toward processing user input and improve immediate interruption handling.

\section{Internal Mechanism of Full-Duplex SLMs}
\subsection{Full-duplex Spoken Language Model}
As shown in Figure~\ref{fig:main}, Full-Duplex Spoken Language Models (FD-SLMs) process two concurrent speech streams: a \emph{user stream} and a \emph{model stream}. 
An audio codec discretizes the continuous speech signals into audio tokens, allowing the interaction to be represented as a sequence of timesteps~\cite{defossez_highfid_neural_audio_comp, zeghidour2021soundstream}. 
At each timestep $t$, the FD-SLM conditions on the incoming user audio tokens and its previously generated model tokens, and then produces the next model response. 
Practically, recent FD-SLMs first generate text tokens as a semantically rich \emph{intermediate representation}, which then guides the generation of the corresponding speech~\cite{chang2026tico, defossez2024moshi, roy2026personaplex, raon_speech_report}.

Formally, at timestep $t$, 
let $u^{(t)}_{\mathrm{audio}}$ denote the user input audio tokens, and let $m^{(t)}_{\mathrm{audio}}$ and $m^{(t)}_{\mathrm{text}}$ denote the model output audio and text tokens, respectively.
Let $M_{\theta}$ denote an FD-SLM parameterized by $\theta$.
At each timestep, $M_{\theta}$ generates the model output tokens $m^{(t)}_{\mathrm{text}}$ and $m^{(t)}_{\mathrm{audio}}$  conditioned on the current user input audio tokens $u^{(t)}_{\mathrm{audio}}$, the model's previous audio and text tokens, and the preceding dialogue context $c^{(t)}$:
\begin{equation}
    \left(m^{(t)}_{\mathrm{audio}}, m^{(t)}_{\mathrm{text}}\right)
    \sim
    M_{\theta}\left(
        \cdot \mid
        u^{(t)}_{\mathrm{audio}},
        m^{(t-1)}_{\mathrm{audio}},
        m^{(t-1)}_{\mathrm{text}},
        c^{(t)}
    \right),
\end{equation}
where $c^{(t)}$ summarizes the dialogue history before timestep $t$.

Throughout the paper, we use a timestep as the minimal unit of processing rather than an individual token. Unlike text-only LLMs, FD-SLMs may contain multiple tokens at each timestep across parallel streams, making timesteps a more consistent unit for our analysis~\cite{defossez2024moshi, aroralandscape, copet2023simple, wu2024codec}.

\subsection{Logit Lens}
\label{sec:logit_lens}
Unlike text-only LLMs or half-duplex SLMs, FD-SLMs must continuously coordinate listening to the user with generation of their own speech. However, how this coordination is represented internally remains poorly understood.
To analyze this internal behavior, we use the \emph{logit lens}~\cite{nostalgebraist2020logitlens, belrose2023eliciting}, which projects hidden representations from intermediate layers into the vocabulary space, allowing us to inspect how token-level predictions evolve across model depth.

Let $h^{(t)} \in \mathbb{R}^{d}$ denote the hidden representation at the selected layer and timestep $t$, and let $W_{\mathrm{unembed}} \in \mathbb{R}^{|V| \times d}$ be the unembedding matrix, where $V$ denotes the token vocabulary. For any target token $y \in V$, we define its projected probability under the hidden representation as
\begin{equation}
\label{eqn:logit_lens_p_def}
P(y \mid h^{(t)}) =
\frac{\exp(w_y^\top h^{(t)})}
{\sum_{v \in V} \exp(w_v^\top h^{(t)})},
\end{equation}
where $w_y^\top$ and $w_v^\top$ are the rows of $W_{\mathrm{unembed}}$ corresponding to tokens $y$ and $v$, respectively.

At each timestep $t$, we then decode the most likely token under this projected distribution:
\begin{equation}
    y_{\text{decode}}^{(t)} = \arg\max_{y \in V} P(y \mid h^{(t)}).
\end{equation}

To understand how the model's internal behavior differs between listening and speaking, we construct a dataset for turn-by-turn interactions, where the model first listens to the user's speech and then speaks to respond. 
We conduct logit-lens analysis on PersonaPlex~\cite{roy2026personaplex} to qualitatively compare hidden-representation predictions between the listening and speaking segments. Further details of the dataset construction are provided in Appendix~\ref{app:dataset_logit_len}.

\begin{finding}[Finding 1]
FD-SLM hidden representations exhibit \textbf{stream-specific predictive focus}: during listening, they preferentially predict the incoming user stream, whereas during speaking, they preferentially predict the output model stream.
\end{finding}

Table~\ref{tab:logit_lens_examples} illustrates the predictive behavior on the user query ``Can you compare renewable energy sources and explain their pros and cons in daily use?'' While the user is speaking, the model stays silent because it is listening. Even so, logit-lens decoding of its intermediate layers anticipates the upcoming user words rather than the model's own output: after hearing ``explain,'' intermediate layers decode tokens such as ``why'' and ``how''; after hearing ``their,'' they decode tokens such as ``own'' and ``pro''; and subsequent predictions align with ``and'' and ``cons.''
During model speaking, in contrast, the decoded tokens track the model's own output stream. Complete layer-wise decoding examples for both segments, together with additional decoded samples, are provided in Appendix~\ref{app:early_decode}.

\begin{table}[htbp]
    \centering
    \caption{Examples of logit-lens decoded predictions during a listening segment. Bold tokens indicate decoded predictions that match or anticipate the actual incoming user speech.}
    \label{tab:logit_lens_examples}
    \renewcommand{\arraystretch}{1.15}
    \fbox{%
    \begin{tabular}{lcccc}
        \toprule
        Current user token & explain & their & pros & and \\
        \midrule
        Intermediate-layer decoded tokens
            & why, how, personal
            & own, \textbf{pro}
            & \textbf{and}
            & \textbf{con}, \textbf{cons} \\
        Actual next user token
            & their
            & \textbf{pros}
            & \textbf{and}
            & \textbf{cons} \\
        \bottomrule
    \end{tabular}%
    }
\end{table}

\subsection{Generative and Perceptive State}
\label{sec:gen_per_state}
The qualitative observation using logit lens suggests that hidden representations exhibit stream-specific predictive focus: their predictions can be more aligned with either incoming user input or model output generation. Building on this observation, we quantify how this predictive focus shifts over time by defining two affinity scores: \emph{generation affinity} and \emph{perception affinity}.

\noindent\textbf{Generation Affinity $\mathcal{S}_{\text{gen}}(t)$:}
Generation affinity $\mathcal{S}_{\text{gen}}(t)$ quantifies the extent to which the hidden representation $h^{(t)}$ supports generation of the output model stream. We define generation affinity as the mean projected probability assigned to the model output text token $m^{(t)}_\mathrm{text}$ and audio token $m^{(t)}_\mathrm{audio}$ conditioned on the current hidden representation $h^{(t)}$:
\begin{equation}
    \mathcal{S}_{\text{gen}}(t) =
    \frac{1}{2}
    \left(
    P(m^{(t)}_\mathrm{audio} \mid h^{(t)})
    +
    P(m_{\text{text}}^{(t)} \mid h^{(t)})
    \right).
    \label{eqn:S_gen}
\end{equation}
A high $\mathcal{S}_{\text{gen}}(t)$ indicates that $h^{(t)}$ is strongly aligned with the model's own output generation, suggesting that the FD-SLM is in a \textit{generative state}.

\noindent\textbf{Perception Affinity $\mathcal{S}_{\text{perc}}(t)$:}
Perception affinity $\mathcal{S}_{\text{perc}}(t)$ quantifies the extent to which the hidden representation $h^{(t)}$ supports prediction of the incoming user stream. We define perception affinity as the projected probability assigned to the next incoming user audio token $u^{(t+1)}_{\mathrm{audio}}$ conditioned on the current hidden representation $h^{(t)}$:
\begin{equation}
    \mathcal{S}_{\text{perc}}(t) =
    P(u_{\mathrm{audio}}^{(t+1)} \mid h^{(t)}).
    \label{eqn:S_perc}
\end{equation}
A high $\mathcal{S}_{\text{perc}}(t)$ indicates that $h^{(t)}$ is strongly aligned with predicting the incoming user audio, suggesting that the FD-SLM is in a \textit{perceptive state}.

We compute $\mathcal{S}_{\text{gen}}(t)$ and $\mathcal{S}_{\text{perc}}(t)$ on the 100 examples from the turn-by-turn interaction dataset. For audio-token probabilities, we use the first codec codebook, which primarily encodes semantic speech information, while later residual codebooks encode finer acoustic details~\cite{defossez2024moshi, zeghidour2021soundstream, defossez_highfid_neural_audio_comp}.\footnote{Using only the first audio codebook also avoids FD-SLM-specific timing offsets associated with later residual codebooks.} We align all examples by setting $t=0$ to the end of the user utterance and average the resulting score trajectories across examples.
For demonstration, we show the results on PersonaPlex.

\begin{finding}[Finding 2]
FD-SLMs coordinate speaking and listening by dynamically modulating between generative and perceptive states.
\end{finding}

\begin{figure}[htbp]
    \centering
    \begin{minipage}{0.48\textwidth}
        \includegraphics[width=\linewidth]{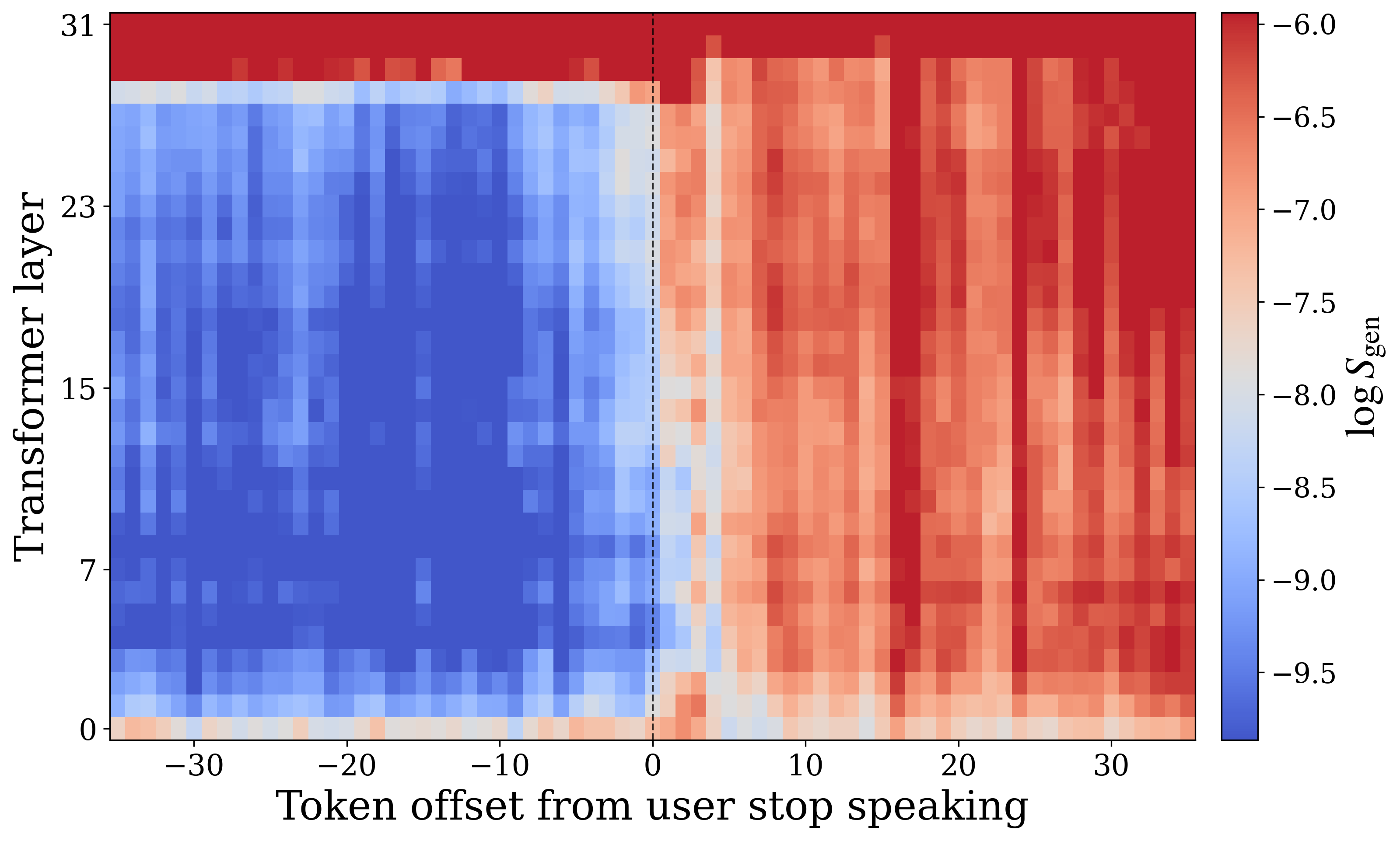}
        \caption{Generation affinity $\mathcal{S}_{\text{gen}}(t)$ across internal layers of PersonaPlex on the turn-by-turn interaction dataset. We align 100 examples at the end of the user utterance, with $t=0$ marking this transition. Values are shown on a logarithmic scale.}
        \label{fig:speaking_LL}
    \end{minipage}\hfill
    \begin{minipage}{0.48\textwidth}
        \includegraphics[width=\linewidth]{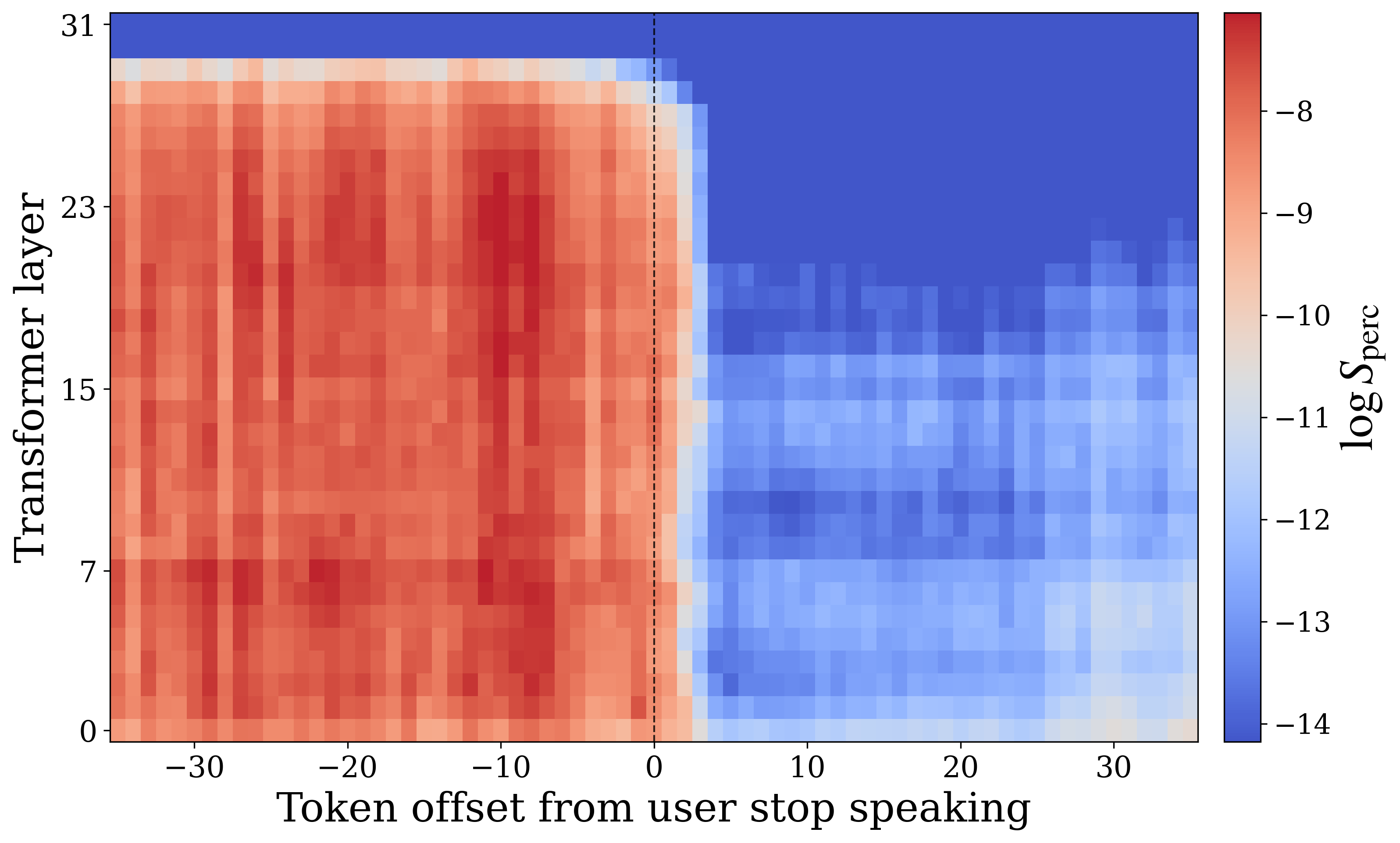}
        \caption{Perception affinity $\mathcal{S}_{\text{perc}}(t)$ across internal layers of PersonaPlex on the turn-by-turn interaction dataset. We align 100 examples at the end of the user utterance, with $t=0$ marking this transition. Values are shown on a logarithmic scale.}
        \label{fig:listening_LL}
    \end{minipage}
\end{figure}

As shown in Figure~\ref{fig:speaking_LL}, $\mathcal{S}_{\text{gen}}(t)$ rises after $t=0$, indicating a transition into the generative state as the model prepares to respond. Conversely, Figure~\ref{fig:listening_LL} shows that $\mathcal{S}_{\text{perc}}(t)$ remains high while the user is speaking ($t<0$), indicating a perceptive state, and then rapidly decays after the user utterance ends. Together, these results show that FD-SLMs do not maintain generation and perception uniformly throughout the interaction; instead, they reconfigure their generative and perceptive states according to the conversational role they currently occupy.

We note that the final layers show a different pattern: $\mathcal{S}_{\text{perc}}(t)$ remains low while $\mathcal{S}_{\text{gen}}(t)$ remains high even during user-speaking segments. This is expected because the final layers are closest to the output distribution and must still produce model tokens at every timestep, which often correspond to silence while the user is speaking. 

\subsection{State Inertia}
\label{sec:state_inertia}

Real-world spoken conversations often involve overlapping speech, including interruptions and backchanneling. Prior work reports that overlap occurs in over 40\% of conversational turns~\cite{lin2026full_v15, heldner2010pauses_and_overlap_speech}, making overlap handling an important capability for FD-SLMs. Unlike half-duplex systems, FD-SLMs are designed to listen while speaking; this simultaneous listening-and-speaking capability is a central motivation for full-duplex speech modeling.

In this work, we focus on user interruption as a representative and practically important form of speech overlapping.
During an interruption, the user begins speaking while the model is still generating, and the model must quickly shift attention to the new input, yield the floor when appropriate, and respond to the updated conversational context. This scenario commonly arises in spoken assistant settings, where users may interrupt system speech to correct an error, redirect the dialogue, or provide input before the system finishes speaking~\cite{strom2000barge_in_in_conversational_systems, raux2008flexible_turn_taking_and_interruption}.

We compare how the generation and perception affinities, $\mathcal{S}_{\text{gen}}(t)$ and $\mathcal{S}_{\text{perc}}(t)$, evolve under two conditions: \textit{interruption} and \textit{no-interruption}.
In the \textit{interruption} condition, we first present a \emph{speech-inducing prompt}: an open-ended question designed to place the model in a generative state. We then interrupt the model using a user utterance from the dataset introduced in the previous section.
In the \textit{no-interruption} condition, we present the same user utterance without first prompting the model to produce a substantive response. 
Detailed dataset construction is presented in Appendix~\ref{app:dataset_state_inertia}
For demonstration, we present an analysis using PersonaPlex as a representative example.

\begin{finding}[Finding 3]
The model exhibits \textbf{state inertia}: a tendency to remain in its prior state even when the conversational context requires an immediate transition.
\end{finding}

\begin{figure}[htbp]
    \centering
    \begin{minipage}{0.48\textwidth}
        \includegraphics[width=\linewidth]{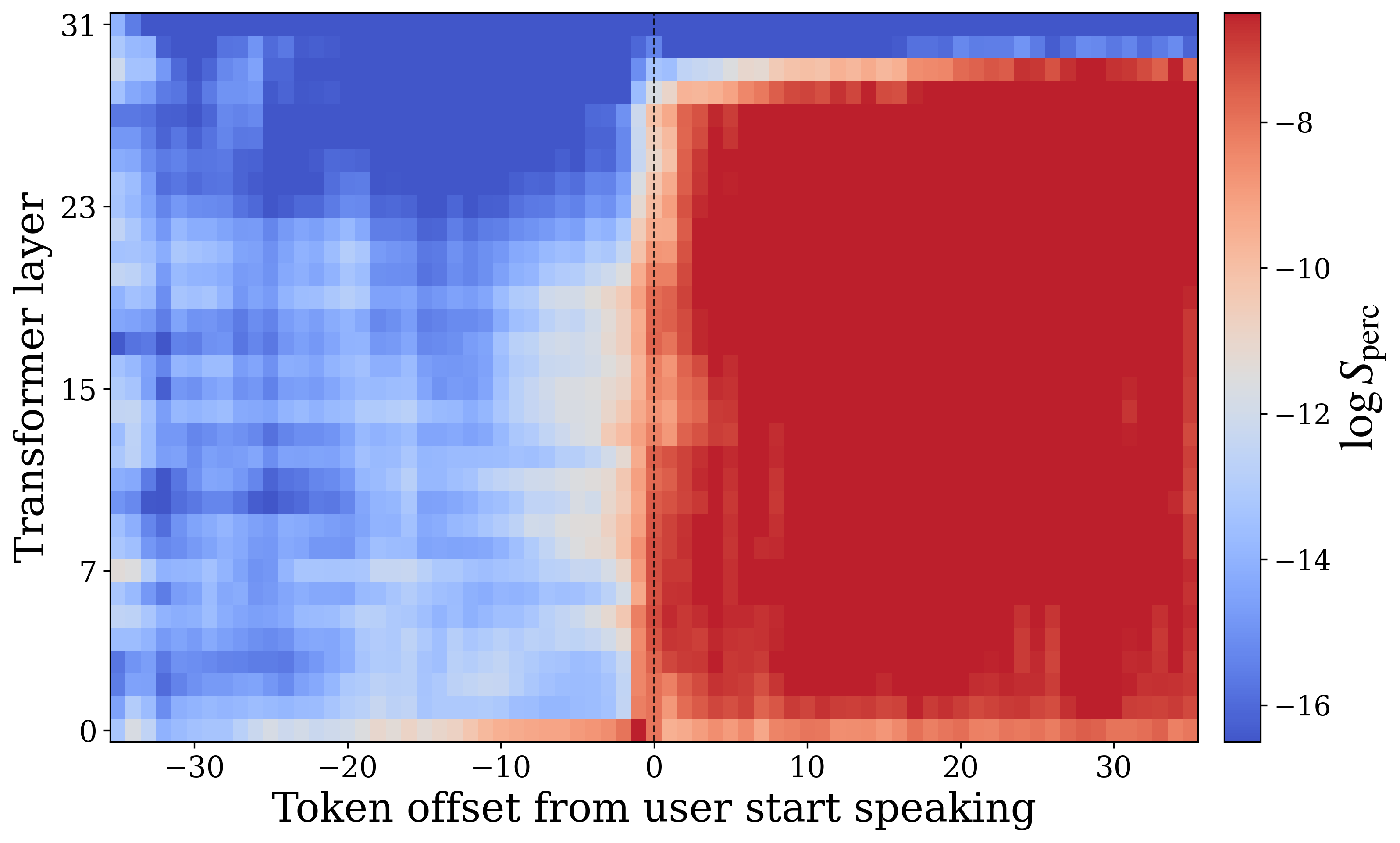}
        \caption{Perception affinity $\mathcal{S}_{\text{perc}}(t)$ in the \textit{no-interruption} condition. The model transitions into the perceptive state immediately after the user begins speaking.}
        \label{fig:S_perc_no_interruption}
    \end{minipage}\hfill
    \begin{minipage}{0.48\textwidth}
        \includegraphics[width=\linewidth]{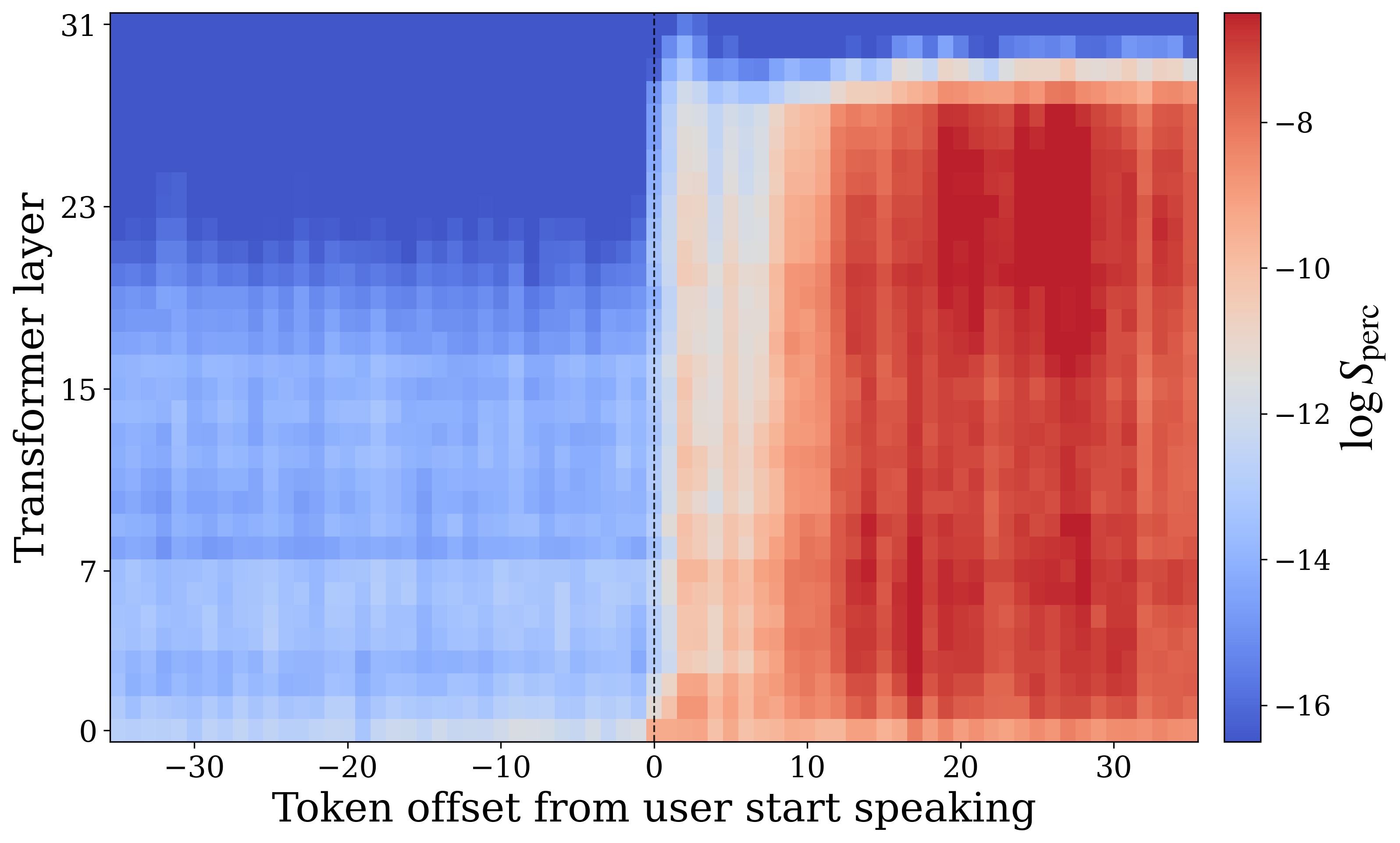}
        \caption{Perception affinity $\mathcal{S}_{\text{perc}}(t)$ in the \textit{interruption} condition. The model transitions into the perceptive state after 7--8 timesteps, exhibiting state inertia.}
        \label{fig:S_perc_interruption}
    \end{minipage}
\end{figure}

As shown in Figures~\ref{fig:S_perc_no_interruption} and~\ref{fig:S_perc_interruption}, $\mathcal{S}_{\text{perc}}(t)$ remains low immediately after abrupt user input in the \textit{interruption} condition compared with the \textit{no-interruption} condition. This indicates that the model does not immediately transition out of the prompt-induced generative state. In this example, $\mathcal{S}_{\text{perc}}(t)$ takes approximately 7--8 timesteps, corresponding to about 0.6 seconds, to recover to the perceptive state. In contrast, under the \textit{no-interruption} condition, the model transitions into the perceptive state almost immediately when the user begins speaking. We observe a similar delay in the generative-state transition, as shown in Appendix~\ref{app:gen_state_inertia}. We refer to this delayed internal transition as \textbf{state inertia}.

\section{Zero-Buffer Benchmark (ZBB)}
\label{sec:zbb_intro}
A question naturally arises:
whether state inertia, the delayed transition into the perceptive state, reduces the model's ability to perceive and understand user interruptions? 
To systematically quantify its impact on dialogue comprehension, we introduce the \emph{Zero-Buffer Benchmark} (ZBB), which evaluates whether FD-SLMs can immediately understand user input when an interruption occurs.
The key design principle is to place the critical semantic content at the very onset of the interrupting utterance, with no leading filler or acoustic buffer, so that the model must perceive core meaning exactly when state inertia is most likely to disrupt perception.

Each ZBB example consists of a \emph{speech-inducing prompt} followed by a \emph{zero-buffer query}. 
The speech-inducing prompt is an open-ended question that places the model in a generative state; while the model is actively responding, we abruptly interrupt it with the zero-buffer query. 
Each zero-buffer query follows the template \texttt{<Subject>, <Description>, <Confirmation Request>} (e.g., \textit{``Submarine flies in the clouds, right?''}), where the subject keyword is deliberately placed as the first word. 
Because the subject carries the information needed to judge the description, missing the onset of the interruption causes the model to lose the subject and often produce an incorrect or incoherent answer.
The detail ZBB dataset creation and examples are provided in Appendix~\ref{app:dataset_zbb}.

For evaluation, we transcribe the generated audio and evaluate the following metrics with an LLM judge:
\begin{itemize}
    \item \textbf{Correctness:} Whether the model answers the zero-buffer query correctly.
    \item \textbf{Initial Word Occurrence Rate (IWOR):} Whether the response explicitly mentions the initial semantic word of the zero-buffer query, or a direct synonym. IWOR provides a diagnostic measure of whether the model perceived the initial subject.
\end{itemize}
Evaluating several recent FD-SLMs on ZBB,
we find that interruption substantially degrades both correctness and IWOR (Section~\ref{sec:zbb_eval_results}), showing that state inertia has a measurable downstream impact on immediate interruption comprehension. 
To address this, the next section introduces a training-free activation steering method that accelerates the model's transition into the perceptive state.



\section{Activation Steering with Perception Vector}
\label{sec:perception_vector}

To mitigate the impact of state inertia, we apply activation steering~\cite{turner2023steering} when the user begins speaking during model generation, shifting the model's hidden representations from the generative state toward the perceptive state.

We classify each timestep $t$ as generation-dominant or perception-dominant using $\mathcal{S}_{\text{gen}}(t)$ and $\mathcal{S}_{\text{perc}}(t)$ computed at intermediate transformer layers. Specifically, we define $T_{\text{gen}} = \{t \mid \mathcal{S}_{\text{gen}}(t) \geq \Theta_{\text{gen}} \wedge \mathcal{S}_{\text{perc}}(t) < \Theta_{\text{perc}}\}$ and $T_{\text{perc}} = \{t \mid \mathcal{S}_{\text{perc}}(t) \geq \Theta_{\text{perc}} \wedge \mathcal{S}_{\text{gen}}(t) < \Theta_{\text{gen}}\}$, where $\Theta_{\text{gen}}$ and $\Theta_{\text{perc}}$ are predefined thresholds.

Following established representation engineering methods~\cite{turner2023steering, zou2023representation, rimsky2024steering}, we construct a \emph{perception vector} as the difference between the mean hidden representations of perception-dominant and generation-dominant timesteps. Let $h^{(t)}$ denote the hidden representation at the selected steering layer and timestep $t$. We define the perception vector $\mu_{g \to p}$, which points from the generative state toward the perceptive state, as
\begin{equation}
    \label{eqn:mean_diff}
    \mu_{g \to p} =
    \frac{1}{|T_{\text{perc}}|}
    \sum_{t \in T_{\text{perc}}} h^{(t)}
    -
    \frac{1}{|T_{\text{gen}}|}
    \sum_{t \in T_{\text{gen}}} h^{(t)}.
\end{equation}

At inference time, we steer the model by adding the perception vector to the hidden representation at the selected steering layer, $\tilde{h}^{(t)} = h^{(t)} + \alpha \mu_{g \to p}$, where $\tilde{h}^{(t)}$ denotes the steered hidden representation and $\alpha$ controls the steering strength. In our ZBB experiments, steering is applied at the onset of the zero-buffer query, with the onset detected by an energy-based detector.

The geometry of the hidden representation space provides additional support for the perception vector. In Appendix~\ref{app:pca}, we show that generation-dominant and perception-dominant timesteps are clearly separated under PCA projection. This separation suggests that the vector captures a meaningful transition direction rather than a noisy difference between overlapping distributions.
\section{Experiments and Results on Zero-Buffer Benchmark}
\label{sec:experiments}

\subsection{Setup}
\label{sec:zbb_setup}

\textbf{Evaluation conditions.}
We evaluate three advanced FD-SLMs spanning distinct architectural paradigms: PersonaPlex~\cite{roy2026personaplex}, Moshi~\cite{defossez2024moshi}, and Raon-SpeechChat~\cite{raon_speech_report}. For each model, we compare three conditions: \textit{no interruption}, \textit{interruption}, and \textit{interruption with steering}. In the \textit{interruption} condition, we first present a speech-inducing prompt and then abruptly interrupt the model with a zero-buffer query. In the \textit{no-interruption} condition, we present the same zero-buffer query without first inducing substantive model speech. This condition represents the model's performance when no generative-to-perceptive transition is required. In the \textit{interruption with steering} condition, we apply the perception vector at the onset of the zero-buffer query and measure whether it restores performance after interruption.

\textbf{Perception vector construction.}
To construct the perception vector, we classify timesteps into $T_{\text{gen}}$ and $T_{\text{perc}}$ using the affinity scores defined in Section~\ref{sec:gen_per_state}. For classification, we average $\mathcal{S}_{\text{gen}}(t)$ and $\mathcal{S}_{\text{perc}}(t)$ over layers 12--24 and apply the thresholds in Table~\ref{tab:steer_param}. Unless otherwise stated, we use the steering layer, steering strength $\alpha$, and steering span $\Delta T_{\text{steer}}$ specified in Table~\ref{tab:steer_param}. Importantly, the conversations used to compute $\mu_{g \to p}$ are drawn from the turn-by-turn interaction dataset introduced in Section~\ref{sec:logit_lens}, and are disjoint from the ZBB evaluation set. Thus, the perception vector captures general state-level differences rather than information specific to the ZBB examples. Representative examples of these conversations are provided in Appendix~\ref{app:dataset}.

\textbf{Steering schedule.}
At inference time, we apply the perception vector $\mu_{g \to p}$ starting at the onset of the zero-buffer query, denoted $t_{\text{int}}$. We detect $t_{\text{int}}$ using an energy-based onset detector. Let $h^{(t)}$ denote the hidden representation at the selected steering layer and timestep $t$. To avoid steering the model throughout the entire interrupted utterance, we apply steering over a finite span $\Delta T_{\text{steer}}$ and linearly decay its magnitude to zero:
\begin{equation}
    \label{eqn:hidden_injection}
    \tilde{h}^{(t)} =
    \begin{cases}
        h^{(t)} +
        \alpha \left(1 - \frac{t - t_{\text{int}}}{\Delta T_{\text{steer}}}\right)
        \mu_{g \to p},
        & t_{\text{int}} \leq t < t_{\text{int}} + \Delta T_{\text{steer}}, \\
        h^{(t)},
        & \text{otherwise},
    \end{cases}
\end{equation}
where $\tilde{h}^{(t)}$ denotes the steered hidden representation and $\alpha$ controls the steering strength.

\subsection{ZBB Evaluation Results}
\label{sec:zbb_eval_results}

As shown in Table~\ref{tab:zbb_multiple_models}, interruption causes a severe degradation in both correctness and IWOR across all three FD-SLMs. On PersonaPlex, for instance, correctness drops from 0.49 to 0.28 and IWOR from 0.74 to 0.40 when the query arrives as an interruption. The IWOR drop in particular indicates that the model often fails to perceive the initial subject of the interrupting utterance, showing that state inertia has a measurable downstream impact on immediate interruption comprehension.

Notably, activation steering improves both correctness and IWOR across all evaluated models. For PersonaPlex and Moshi, the perception vector raises response correctness and restores most of the interruption-induced IWOR drop (94\% and 92\%, respectively). For Raon-SpeechChat, steering improves both metrics as well, though absolute correctness remains low.

\begin{figure}[htbp]
    \centering
    \begin{minipage}{0.48\textwidth}
        \includegraphics[width=\linewidth]{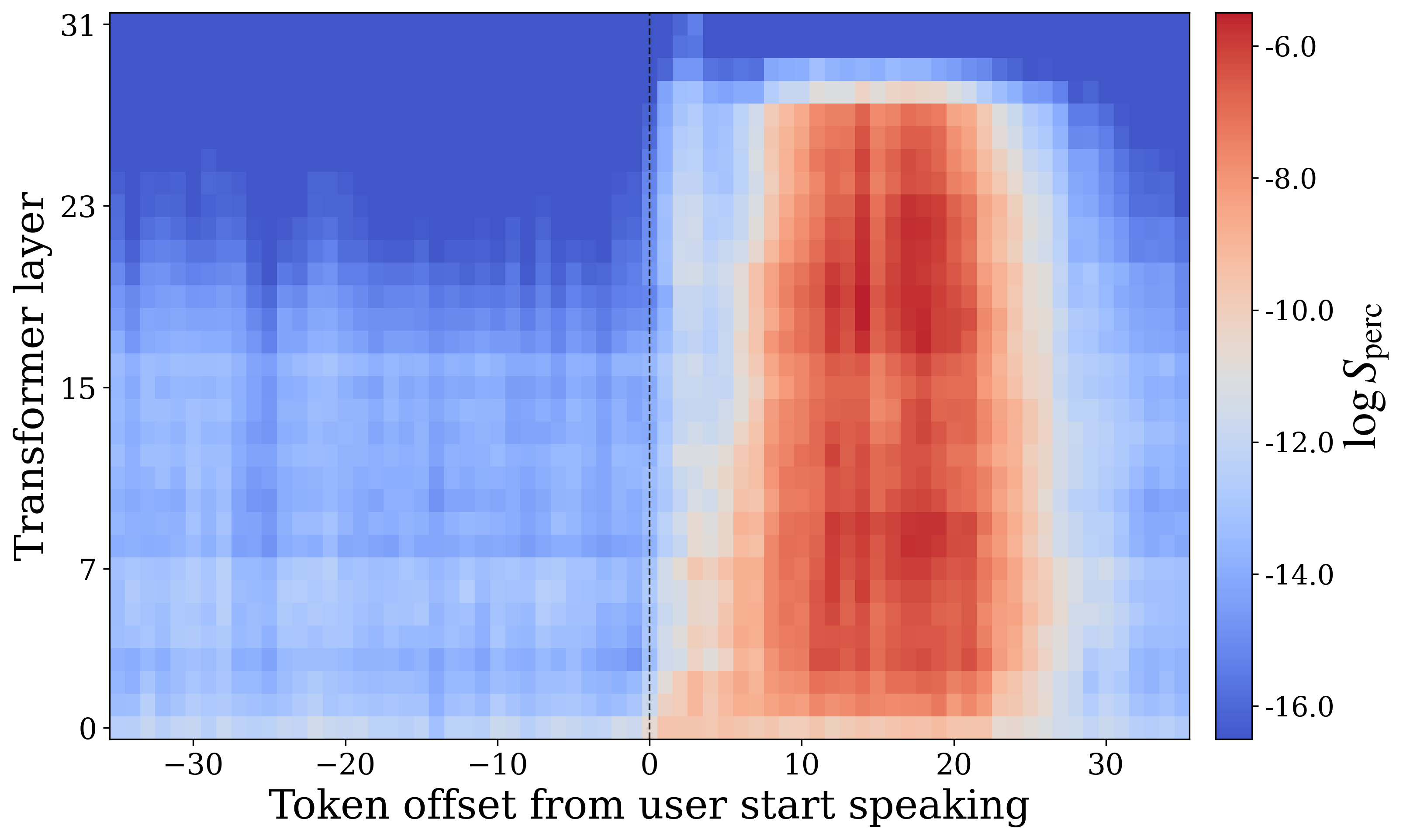}
        \caption{Perception affinity $\mathcal{S}_{\text{perc}}(t)$ in the \textit{interruption} condition. Without steering, perception affinity takes approximately 7--8 timesteps to recover after interruption, indicating state inertia.}
        \label{fig:no_steer_listen_LL}
    \end{minipage}\hfill
    \begin{minipage}{0.48\textwidth}
        \includegraphics[width=\linewidth]{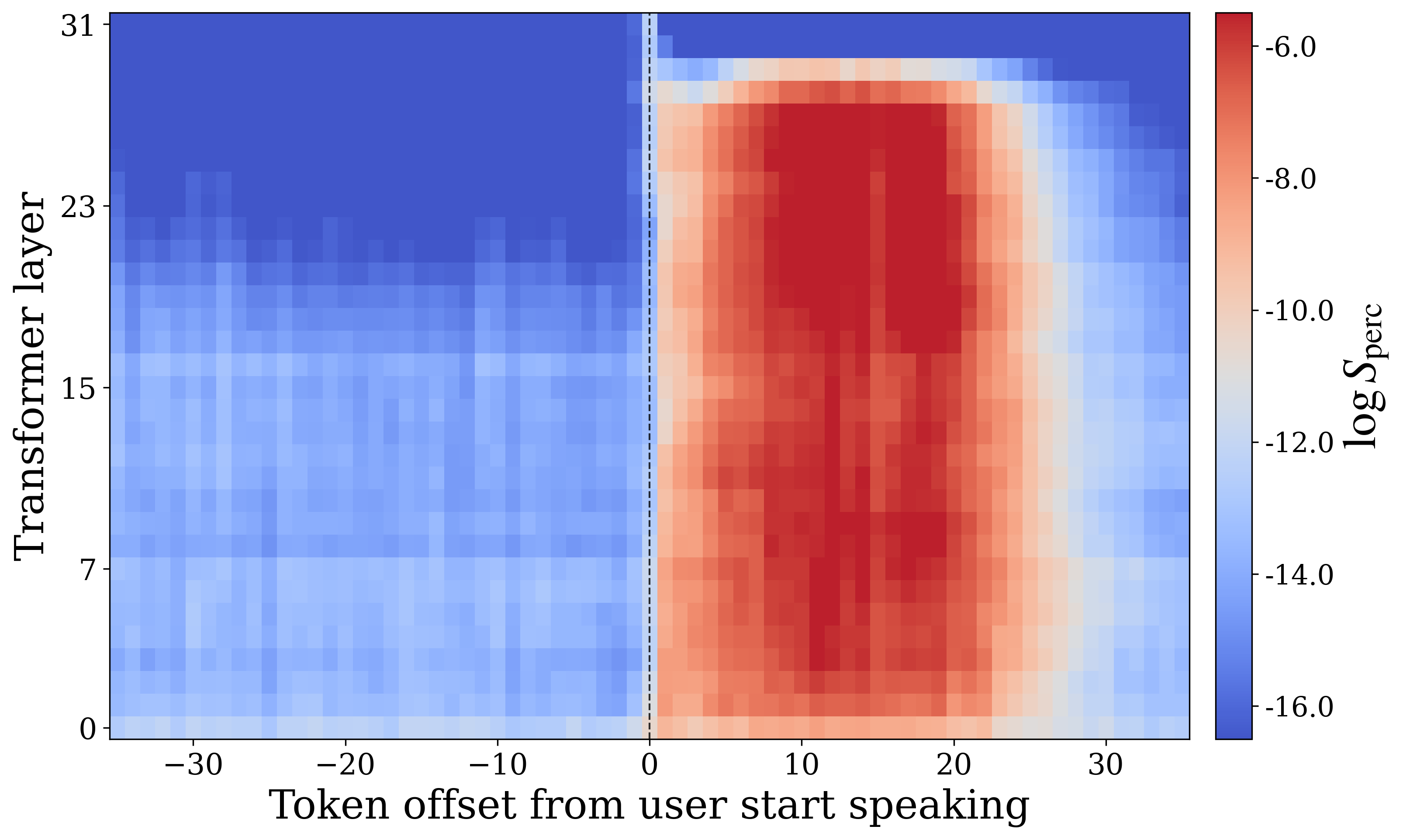}
        \caption{Perception affinity $\mathcal{S}_{\text{perc}}(t)$ in the \textit{interruption with steering} condition. With activation steering, perception affinity recovers immediately after interruption, indicating a faster transition toward the perceptive state.}
        \label{fig:steer_listen_LL}
    \end{minipage}
\end{figure}

We further show qualitatively that activation steering reduces state inertia.
We compare $\mathcal{S}_{\text{perc}}(t)$ around the onset of the zero-buffer query under the \textit{interruption} and \textit{interruption with steering} conditions in Figures~\ref{fig:no_steer_listen_LL} and~\ref{fig:steer_listen_LL}, respectively. In the \textit{interruption} condition, $\mathcal{S}_{\text{perc}}(t)$ remains low immediately after the zero-buffer query begins, indicating a delayed transition into the perceptive state. In contrast, under \textit{interruption with steering}, $\mathcal{S}_{\text{perc}}(t)$ recovers immediately after the zero-buffer query onset. We provide an attention-based analysis in Appendix~\ref{app:attention_recovery}, showing that steering increases attention to the first few interruption timesteps. Additional steering-parameter sweeps are provided in Appendix~\ref{app:steering_param_sweep}.

We also evaluate steering on Full-Duplex Bench (FDB)~\cite{lin2025full_v1} and confirm that steering does not degrade overall full-duplex dialogue performance. Results and discussion are provided in Appendix~\ref{app:fdb_results}.


\begin{table}[t]
\centering
\small

\begin{minipage}{0.60\linewidth}
\centering
\caption{
FD-SLMs performance on ZBB. Uncertainties denote one standard error; parentheses show the percentage of the interruption-induced drop recovered by steering.
}
\label{tab:zbb_multiple_models}
\setlength{\tabcolsep}{5pt}
\renewcommand{\arraystretch}{1.08}
\resizebox{\linewidth}{!}{
\begin{tabular}{@{}llcc@{}}
\toprule
Model & Scenario & Correctness& IWOR \\
\midrule
\multirow{3}{*}{PersonaPlex}
    & No Interrupt    & $0.49 \pm 0.05$ & $0.74 \pm 0.04$ \\
    & Interrupt       & $0.28 \pm 0.04$ & $0.40 \pm 0.05$ \\
    & Interrupt+Steer & $\mathbf{0.45 \pm 0.05}$ {\textcolor{ForestGreen}{(81\%)}}& $\mathbf{0.72 \pm 0.04}$  {\textcolor{ForestGreen}{(94\%)}}\\
\midrule
\multirow{3}{*}{Moshi}
    & No Interrupt    & $0.43 \pm 0.05$ & $0.67 \pm 0.05$ \\
    & Interrupt       & $0.22 \pm 0.04$ & $0.29 \pm 0.05$ \\
    & Interrupt+Steer & $\mathbf{0.34 \pm 0.05}$ {\textcolor{ForestGreen}{(57\%)}} & $\mathbf{0.64 \pm 0.05}$ {\textcolor{ForestGreen}{(92\%)}} \\
\midrule
\multirow{3}{*}{Raon}
    & No Interrupt    & $0.10 \pm 0.03$ & $0.29 \pm 0.05$ \\
    & Interrupt       & $0.03 \pm 0.02$ & $0.16 \pm 0.04$ \\
    & Interrupt+Steer & $\mathbf{0.17 \pm 0.03}$ {\textcolor{ForestGreen}{(200\%)}} & $\mathbf{0.24 \pm 0.04}$ {\textcolor{ForestGreen}{(62\%)}} \\
\bottomrule
\end{tabular}
}
\end{minipage}
\hfill
\begin{minipage}{0.37\linewidth}
\centering
\caption{Activation steering hyperparameters. Thresholds are reported in natural-log scale.}
\label{tab:steer_param}
\setlength{\tabcolsep}{4pt}
\renewcommand{\arraystretch}{1.08}
\resizebox{\linewidth}{!}{
\begin{tabular}{@{}lccc@{}}
\toprule
 & PersonaPlex & Moshi & Raon \\
\midrule
Layer & 23 & 23 & 26 \\
$\alpha$ & 5.5 & 5.5 & 1.2 \\
$\Delta T_{\text{steer}}$ & 3 & 3 & 3 \\
$\ln \Theta_{\text{gen}}$ & -3.5 & -3.5 & -7.5 \\
$\ln \Theta_{\text{perc}}$ & -3.9 & -3.9 & -9.5 \\
\bottomrule
\end{tabular}
}
\end{minipage}

\end{table}

\section{Limitations}
\label{sec:limitations}
Our work has several limitations. First, the steering method relies on detecting the onset of user interruption. We use an energy-based onset detector, but real-world deployment may require more robust voice activity detection, especially in noisy or multi-speaker settings. We discuss false-trigger sensitivity in Appendix~\ref{app:false_trigger}. Second, our evaluation is constrained by the limited availability of open-source FD-SLMs, as few such models are currently publicly available. Finally, our logit-lens-based affinity scores are diagnostic approximations and can be noisy for individual examples.

\section{Conclusion}
\label{sec:conclusion}
We study how FD-SLMs coordinate listening and speaking through hidden representations. Using logit-lens-based affinity scores, we find that FD-SLMs exhibit stream-specific predictive focus and modulate between generative and perceptive states. We identify \emph{state inertia}, a delayed transition during abrupt interruptions that causes models to miss early user input. To evaluate this failure mode, we introduce the Zero-Buffer Benchmark (ZBB) and show that interruption degrades both correctness and IWOR across multiple FD-SLMs. Finally, activation steering with the perception vector reduces state inertia and improves interruption handling without fine-tuning. Overall, our results show that hidden representations can be used not only to analyze FD-SLM listening--speaking coordination, but also to improve full-duplex interruption robustness.

\bibliographystyle{plainnat}
\bibliography{references}

\clearpage
\appendix
\section{Dataset Details}
\label{app:dataset}

\subsection{Turn-by-turn interaction dataset}
\label{app:dataset_logit_len}
\begin{figure}[htbp]
    \centering
    \includegraphics[
        page=1,
        width=0.8\textwidth
    ]{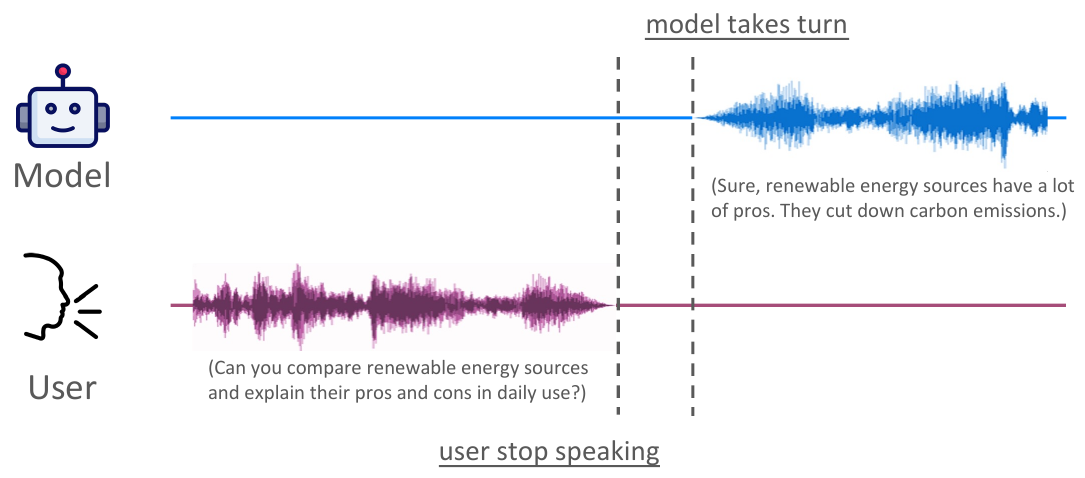}
    \caption{An example from the turn-by-turn interaction dataset used for logit-lens analysis and model-internal generation/perception affinity analysis.}
    \label{fig:A1-dataset}
\end{figure}
The turn-by-turn interaction dataset consists of 100 \emph{user queries} covering a diverse set of everyday conversational topics, each followed by a response window in which the model takes its turn to reply. 
We use this dataset for our logit-lens analysis, and to identify the generative and perceptive states by computing the generation and perception affinities.

We generate these user queries with a text-based LLM (Claude Opus 4.5) according to the following criteria: (1) the utterances should cover varied topics from daily conversation in order to increase diversity; (2) they should be open-ended, so that model responses are not biased toward a fixed answer format; and (3) after text-to-speech synthesis, they should correspond to approximately 15--20 seconds of speech, providing a sufficiently long listening segment for analysis. Example queries are shown below.

\begin{finding}[Example User query 1]
Can you compare renewable energy sources and explain their pros and cons in daily use?
\end{finding}

\begin{finding}[Example User query 2]
My neighbor got this new puppy last week. Cutest little thing you've ever seen, but it barks all night long. I mean, non-stop. I haven't slept properly in days. I don't want to be rude about it, but I'm seriously considering saying something to her about the noise.
\end{finding}

After generating the text queries, we synthesize them into speech using the \texttt{Dia2-2B} text-to-speech (TTS) model\footnote{\url{https://huggingface.co/nari-labs/Dia2-2B}}. 
Because FD-SLMs operate on continuous audio input, each synthesized user utterance is followed by a 10-second silence segment, during which the model is allowed to respond. Thus, each audio input is approximately 25--30 seconds long: the first 15--20 seconds contain user speech, during which the model is expected to listen, and the final 10 seconds provide a response window for the model. The dataset contains 100 such examples.

\subsection{Interruption and No-Interruption Conditions for Analyzing State Inertia}
\label{app:dataset_state_inertia}
\begin{figure}[htbp]
    \centering
    \includegraphics[
        page=1,
        width=\textwidth
    ]{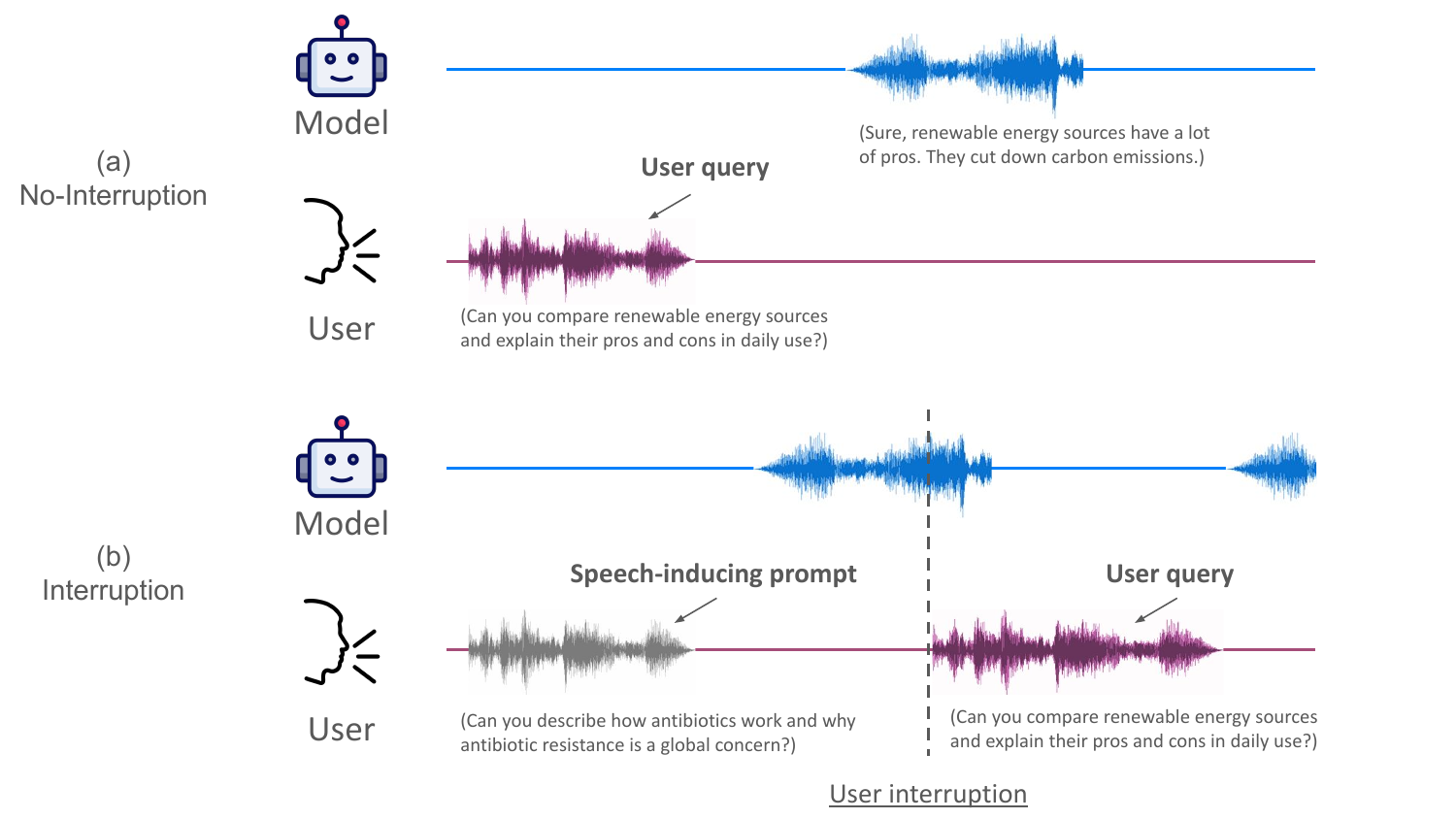}
    \caption{An example from the dataset for state inertia analysis, illustrating the paired (a) no-interruption and (b) interruption conditions. In the interruption condition, a speech-inducing prompt first places the model in a generative state, and a user utterance then interrupts its ongoing response; in the no-interruption condition, the same utterance is presented without a preceding prompt.}
    \label{fig:A2-dataset}
\end{figure}

To analyze state inertia, we construct paired \emph{no-interruption} and \emph{interruption} conditions from the same user queries in Appendix~\ref{app:dataset_logit_len}.

For the no-interruption condition, we present a user query on its own. The model is therefore not speaking when the user begins, yielding an ordinary turn-taking dialogue with no overlap. This setting is the same as in the turn-by-turn interaction dataset.

For the interruption condition, we first input a user \emph{speech-inducing prompt}, which is an open-ended question designed to drive the model into a sustained generative state by eliciting a long response. 
These speech-inducing prompts are constructed according to the following criteria:
(1) they should cover diverse topics to reduce topic bias; 
(2) they should involve relatively technical or explanatory content, so that the model is likely to produce a longer response; and 
(3) they do not need to be long, since their purpose is only to induce model-side speaking behavior. The speech-inducing prompts are generated using Claude Opus 4.5 and synthesized into speech using \texttt{Dia2-2B}.

An example speech-inducing prompt is shown below.
\begin{finding}[Example Speech-Inducing Prompt]
Can you describe how antibiotics work and why antibiotic resistance is a global concern?
\end{finding}

After receiving the speech-inducing prompt, the model begins generating a response; 
after 5 seconds, we abruptly interrupt it with the user query. This setup creates an interruption condition in which the model must transition from an ongoing generative state to a perceptive state.

\subsection{Zero-Buffer Benchmark Dataset}
\label{app:dataset_zbb}
\begin{figure}[htbp]
    \centering
    \includegraphics[
        page=1,
        width=\textwidth
    ]{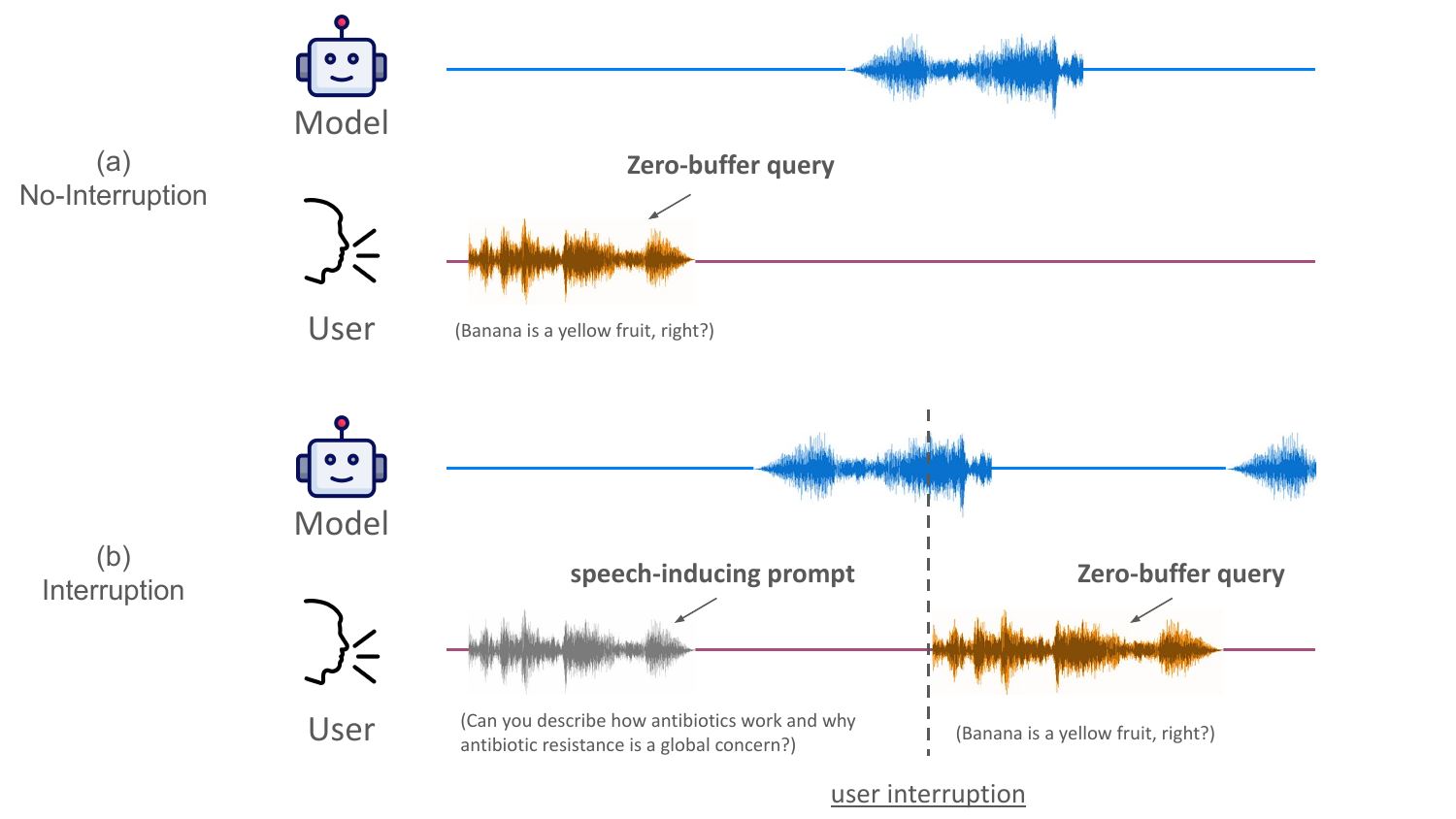}
    \caption{An example from the ZBB dataset, showing the paired (a) no-interruption and (b) interruption conditions. In the no-interruption condition, the zero-buffer query is presented on its own. In the interruption condition, a speech-inducing prompt is followed by a zero-buffer query that interrupts the model's ongoing response, testing whether the model can perceive the critical information at the onset of the interruption.}
    \label{fig:A3-dataset}
\end{figure}

As described in Section~\ref{sec:zbb_intro}, the Zero-Buffer Benchmark (ZBB) contains two evaluation conditions: an interruption condition and a no-interruption condition. In the interruption condition, each example consists of a \emph{speech-inducing prompt} followed by a \emph{zero-buffer query}. 
In the no-interruption condition, the model receives the same zero-buffer query without first being induced into a sustained speaking state. This paired design allows us to measure how interruption affects both response correctness and initial-word recognition.

The speech-inducing prompts are constructed in the same way as in Appendix~\ref{app:dataset_state_inertia}. Each zero-buffer query follows the template
\begin{center}
\texttt{<Subject>, <Description>, <Confirmation Request>}.
\end{center}
The subject appears as the first word of the query, so missing the onset of the interruption often removes the key information needed to answer correctly. To balance the dataset, we generate 50 subjects. For each subject, we create one factually correct description and one factually incorrect description, resulting in 100 zero-buffer queries in total. The confirmation request is kept short, so that the first word remains the primary semantic cue at the onset of the interruption.

The subjects are chosen from common entities, objects, and animals, so that the expected answer is unambiguous and does not require specialized knowledge.

An example positive--negative pair is shown below.

\begin{finding}[Example Zero-Buffer Query Pair]
\textbf{Correct:} Banana is a yellow fruit, right? \\[2pt]
\textbf{Incorrect:} Banana is a red fruit, right?
\end{finding}

Pairing the same subject with both a correct and an incorrect description helps control for subject-specific difficulty. In this way, differences in correctness are less likely to be explained by some subjects being inherently easier or harder to recognize.

The speech-inducing prompts and zero-buffer queries are synthesized into audio using the \texttt{Dia2-2B} text-to-speech model\footnote{\url{https://huggingface.co/nari-labs/Dia2-2B}}.

\subsection{LLM-Based Evaluation for ZBB}
\label{app:zbb_llm_eval}

We evaluate model responses using two metrics: correctness and Initial Word Occurrence Rate (IWOR). For both metrics, we first transcribe the model's generated speech into text using the ASR model \texttt{nvidia/parakeet-tdt-0.6b-v2}\footnote{\url{https://huggingface.co/nvidia/parakeet-tdt-0.6b-v2}}. We then evaluate the transcription using GPT-4.1-mini with the prompts below.

For correctness, the evaluator determines whether the model gives a factually correct and direct answer to the interruption query.

\begin{promptbox}[CORRECTNESS\_SYSTEM\_PROMPT]
Evaluate if the AI provided a FACTUALLY CORRECT and DIRECT answer to the
User's Interruption Question. Do NOT evaluate politeness, tone, or
conversational flow. Be strictly objective about accuracy.

Score 1: The response provides an accurate, specific, and logically correct
answer to the interruption question.
Score 0: The response is incorrect, evasive, or answers the wrong question.

Automatically assign 0 for any of these failures:
- Topic Bleed (Inertia): The AI continues answering or referencing the
  pre-interruption topic, failing to fully transition its logic to the
  new question.
- Factually Incorrect/Hallucination: The AI attempts to answer the new
  question but provides definitively wrong information.
- Evasion/Refusal: The AI dodges the question or gives a meaningless
  generic response.
- Vague Approximation: The answer lacks the specific detail required by
  the question.

Output ONLY JSON:
{"score": <0 or 1>, "reason": "<1-sentence explanation stating what the
correct answer should be and why the AI passed/failed>"}
\end{promptbox}

For IWOR, the evaluator determines whether the model response explicitly mentions the subject entity appearing as the first word of the interruption query, or a direct synonym. This metric is designed to measure whether the model perceived the initial semantic keyword of the interruption.

\begin{promptbox}[FIRST\_WORD\_SYSTEM\_PROMPT]
Evaluate if the AI successfully heard the VERY FIRST WORD of the
User's Interruption Question. Do NOT infer context. Be strictly literal.

Score 1: The response EXPLICITLY names the entity/subject of the first
word (or a direct synonym).
Score 0: The response does NOT explicitly name the first word's subject.

Automatically assign 0 for any of these failures:
- Tail-End Catching: Reacts only to the end of the question, missing the
  first word.
- Pronoun Dodging: Uses "it/they/that/this" instead of explicitly naming
  the subject.
- Self-Referential: Answers with "I/my/me" because it missed the subject.
- Naked Answers: Just says "Yes/No/True/False" with no subject attached.
- Unrelated/Gibberish: Fails to address the first word entirely.

Output ONLY JSON:
{"score": <0 or 1>, "reason": "<1-sentence explanation identifying the
first word and whether it was explicitly said>"}
\end{promptbox}

The final correctness score is the fraction of examples for which the evaluator assigns a score of 1 under the correctness rubric. The final IWOR score is the fraction of examples for which the evaluator assigns a score of 1 under the first-word rubric.

The following example illustrates the correctness evaluation.

\begin{promptbox}[Example of Correctness Evaluation]
{
  "interruption_question": "Bicycle has four wheels, right?",
  "model_response": "Yeah, a bicycle has four wheels, that's right.",
  "gpt_score": 0,
  "gpt_reason": "The response incorrectly states that a bicycle has four wheels, whereas a bicycle actually has two wheels."
}
\end{promptbox}

The following example illustrates the IWOR evaluation.

\begin{promptbox}[Example of IWOR Evaluation]
{
  "interruption_question": "Bicycle has two wheels, right?",
  "model_response": "Yeah, that's right. A bike has two wheels.",
  "gpt_score": 1,
  "gpt_reason": "The first word 'Bicycle' is explicitly referred to as 'A bike' in the response."
}
\end{promptbox}

Correctness and IWOR capture complementary aspects of interruption handling. Correctness measures whether the model answers the full interruption query accurately, whereas IWOR measures whether the model perceived the initial semantic keyword. A model may answer incorrectly even after recognizing the first word, or it may respond to the tail end of the question without explicitly recognizing the subject. We therefore report both metrics.
\section{Computational Resources}
\label{app:compute}

All experiments in this paper are conducted on NVIDIA L40S GPUs. Our experiments involve inference-time analysis and activation steering on open-source FD-SLMs, without model training or fine-tuning. Therefore, the compute requirements are modest compared with training-based approaches. The experiments can be run on any GPU with sufficient memory to host the evaluated models, including PersonaPlex, Moshi, and Raon-SpeechChat.
\section{Delayed Transition Out of the Generative State}
\label{app:gen_state_inertia}

\begin{figure}[htbp]
    \centering
    \begin{minipage}{0.48\textwidth}
        \includegraphics[width=\linewidth]{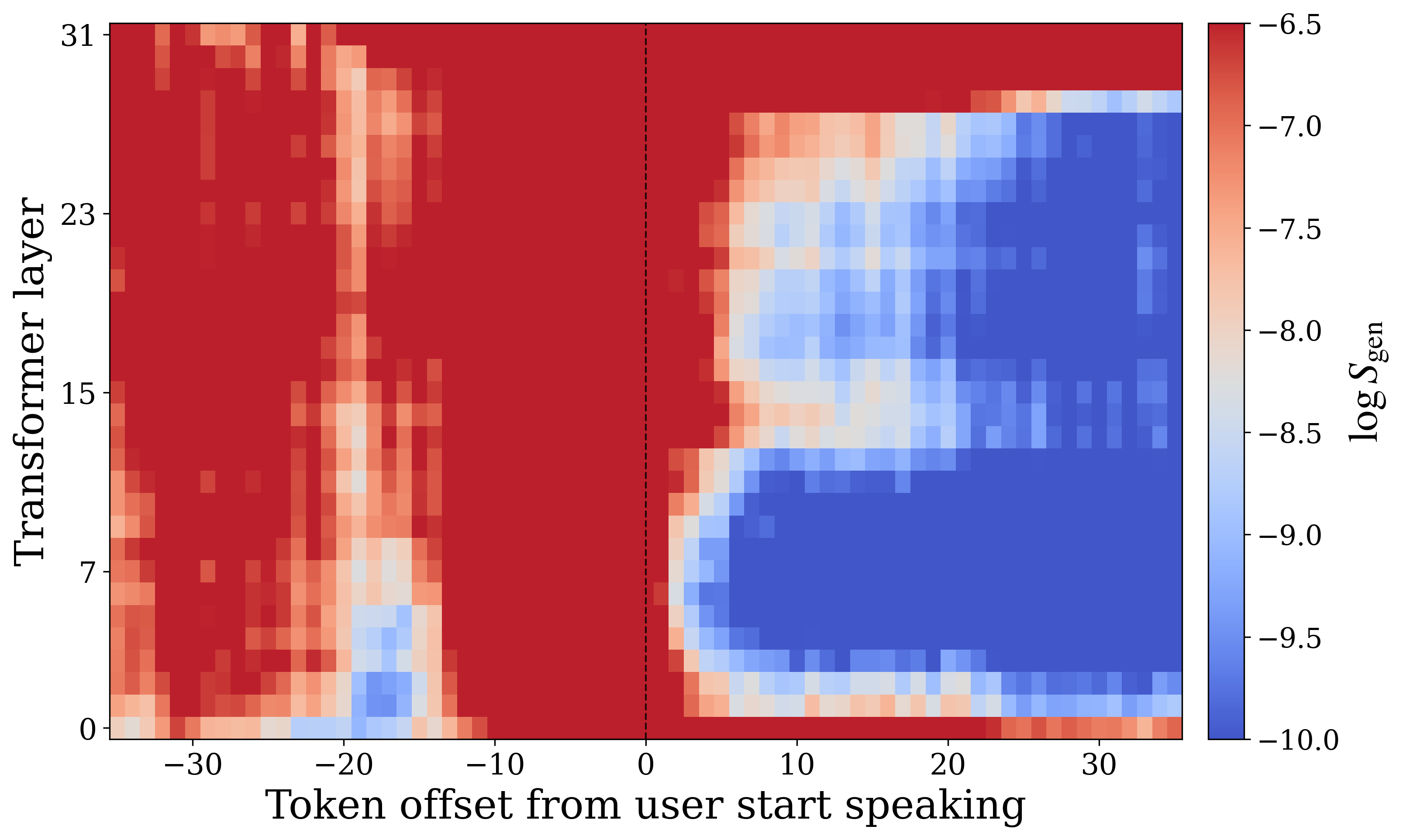}
        \caption{Generation affinity $\mathcal{S}_{\text{gen}}(t)$ in the \textit{no-interruption} condition. The model exits the generative state soon after the user begins speaking, with recovery occurring after approximately 5 timesteps.}
        \label{fig:S_gen_no_interruption}
    \end{minipage}\hfill
    \begin{minipage}{0.48\textwidth}
        \includegraphics[width=\linewidth]{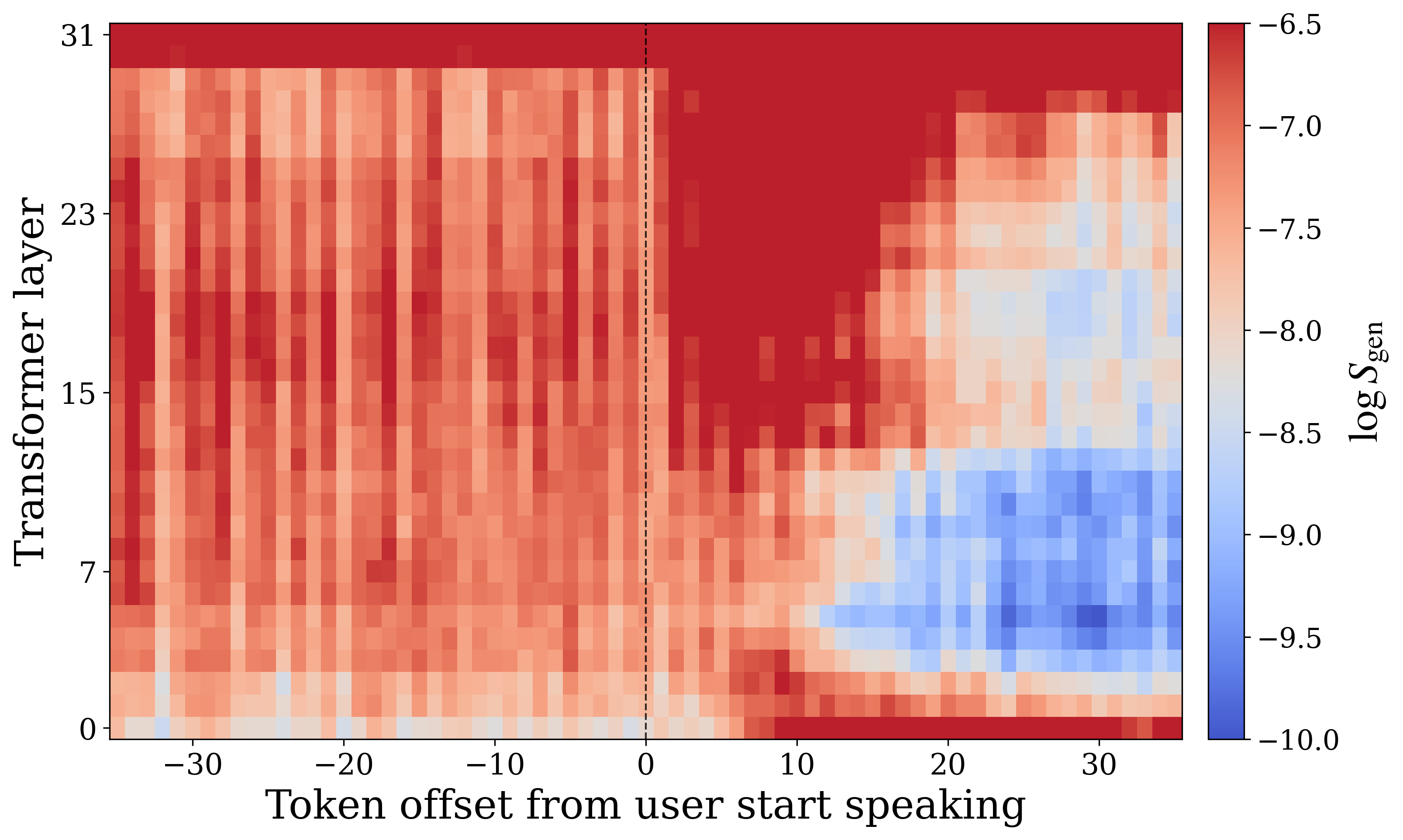}
        \caption{Generation affinity $\mathcal{S}_{\text{gen}}(t)$ in the \textit{interruption} condition. The model remains in the generative state for approximately 20 timesteps after the user interrupts and begins speaking, corresponding to nearly 2 seconds.}
        \label{fig:S_gen_interruption}
    \end{minipage}
\end{figure}

In addition to the delayed transition into the perceptive state discussed in the main text, we also observe a delayed transition out of the generative state. Figure~\ref{fig:S_gen_no_interruption} and Figure~\ref{fig:S_gen_interruption} compare $\mathcal{S}_{\text{gen}}(t)$ under the \textit{no-interruption} and \textit{interruption} conditions, respectively. Under the \textit{no-interruption} condition, generation affinity decreases shortly after the user begins speaking, indicating that the model exits the generative state relatively quickly. In contrast, under the \textit{interruption} condition, $\mathcal{S}_{\text{gen}}(t)$ remains elevated for substantially longer after the user begins speaking, indicating that the model continues to occupy the generative state despite the change in conversational context. This provides complementary evidence for \textbf{state inertia}: the model exhibits a delayed internal transition not only into the perceptive state, but also out of the generative state.

\section{PCA of Hidden Representations}
\label{app:pca}

The perception vector $\mu_{g \to p}$ is computed as the difference between the mean hidden representations of perception-dominant and generation-dominant timesteps. This mean-difference direction is meaningful only if the two underlying representation distributions are sufficiently separated; if they heavily overlap, the resulting vector could instead reflect noise from weakly distinguishable distributions. To examine this possibility, we analyze the separability of these hidden representations using Principal Component Analysis (PCA).

\begin{figure}[htbp]
    \centering
    \includegraphics[width=0.98\linewidth]{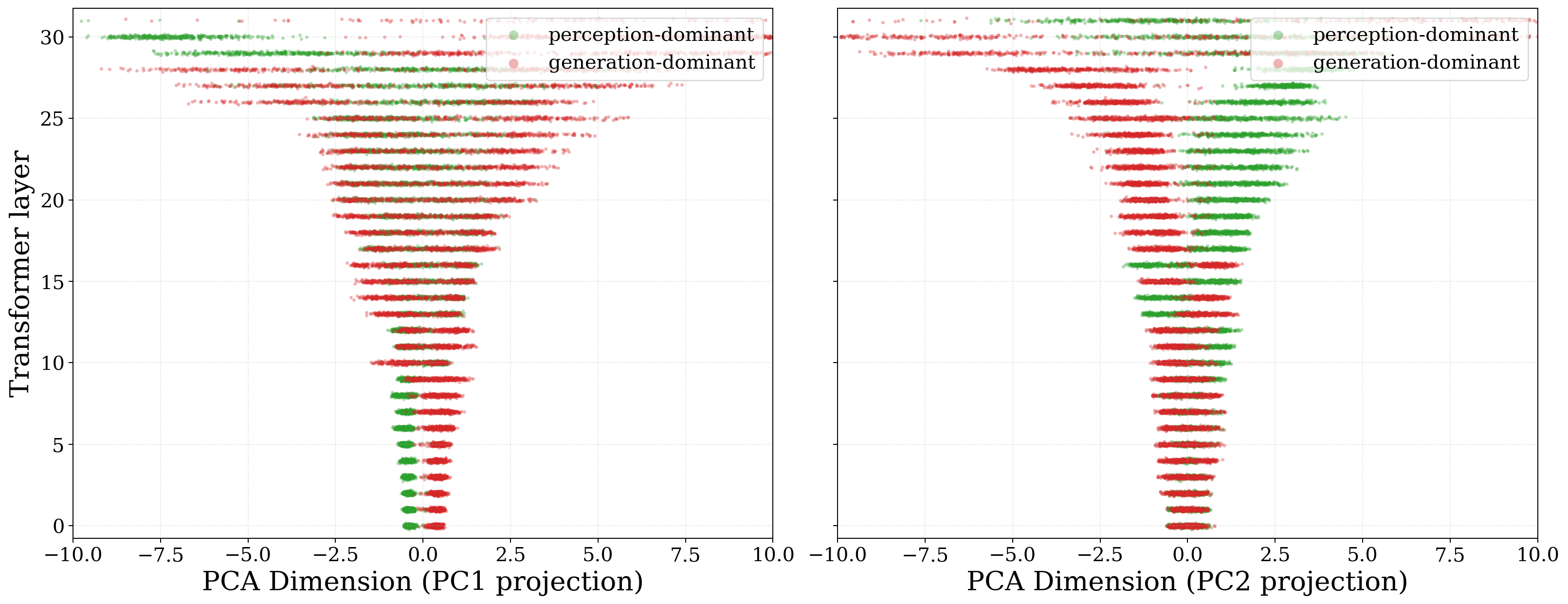}
    \caption{PCA projections of hidden representations from generation-dominant and perception-dominant timesteps across transformer layers. Generation-dominant and perception-dominant representations form separated clusters in the projected space. The separation is most visible along the first principal component in shallower layers (left) and along the second principal component in deeper layers (right).
    }
    \label{fig:pca}
\end{figure}

As shown in Figure~\ref{fig:pca}, generation-dominant and perception-dominant timesteps form clearly separated clusters in the PCA-projected hidden space across most layers. This separation supports the validity of the perception vector: it is not merely a noisy difference between overlapping distributions, but a direction aligned with a prominent structure in the model's hidden representations.

The dominant separating component varies across depth. In lower layers, the two sets are primarily separated along the first principal component, whereas in deeper layers the separation becomes more apparent along the second principal component. One possible interpretation is that the dominant sources of variance change across layers: lower layers may emphasize surface-level or modality-specific structure, while deeper layers may allocate the leading principal component to content-related variation~\cite{geva2021transformer, tenney2019bert}, leaving state-related variation to appear in a secondary component. We treat this explanation as suggestive rather than conclusive.

\section{Decoding Hidden States with the Logit Lens}
\label{app:early_decode}

This appendix provides detailed qualitative examples from the turn-by-turn interaction dataset, complementing the analysis in Section~\ref{sec:logit_lens}. We visualize the top logit-lens prediction at each layer and timestep. For each hidden representation $h^{(t)}$, we project it into the vocabulary space using the same probability definition as in Section~\ref{sec:logit_lens}, and decode
\begin{equation}
y_{\mathrm{decode}}^{(t)}
=
\arg\max_{y \in V} P(y \mid h^{(t)}).
\end{equation}
In each heatmap, the text annotation in a cell shows $y_{\mathrm{decode}}^{(t)}$, while the color indicates the projected probability assigned to the eventual model-side text token $m_{\mathrm{text}}^{(t)}$.

\begin{table}[htbp]
    \centering
    \caption{Examples of logit-lens decoded predictions during listening. Bold tokens match or anticipate the actual upcoming user-side token.}
    \label{tab:logit_lens_examples_appendix}
    \renewcommand{\arraystretch}{1.15}
    \fbox{%
    \begin{tabular}{lll}
        \toprule
        Current user token & Intermediate-layer decoded tokens & Actual next user token \\
        \midrule
        explain & why, how, personal & their \\
        their   & own, \textbf{pro} & \textbf{pros} \\
        pros    & \textbf{and} & \textbf{and} \\
        and     & \textbf{con}, \textbf{cons} & \textbf{cons} \\
        \bottomrule
    \end{tabular}%
    }
\end{table}

\subsection{Logit-Lens Decoding During Listening}

\begin{figure}[htbp]
    \centering
    \includegraphics[width=\linewidth]{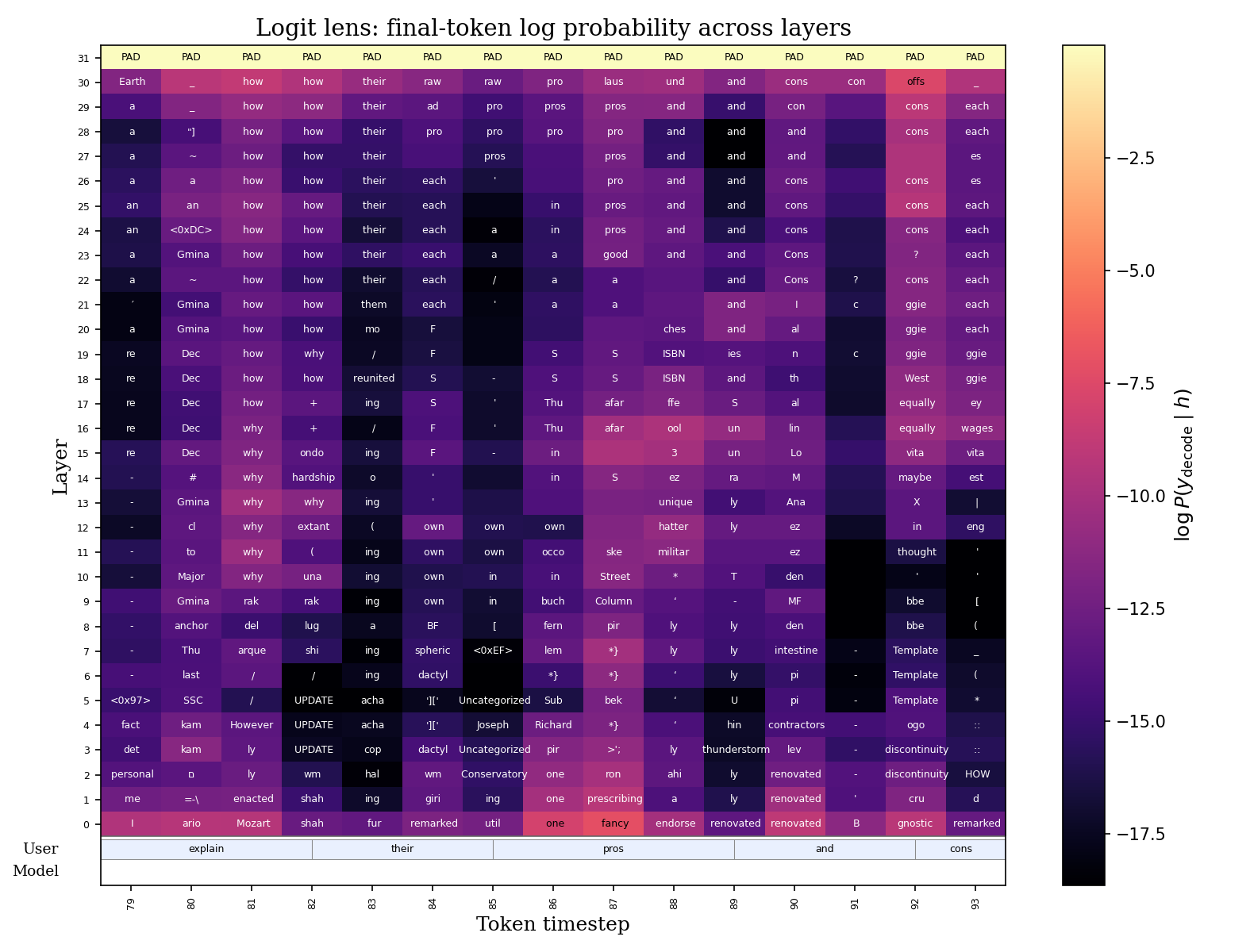}
    \caption{Logit-lens decoding of PersonaPlex hidden states during a listening segment. Intermediate layers often decode tokens related to the incoming user stream, even though the final model-side output remains mostly \texttt{<PAD>}. This suggests that the model internally tracks user-side content before converting this computation into a silent model-side output.}
    \label{fig:logit_len_decode_listen}
\end{figure}

\begin{figure}[htbp]
    \centering
    \includegraphics[width=\linewidth]{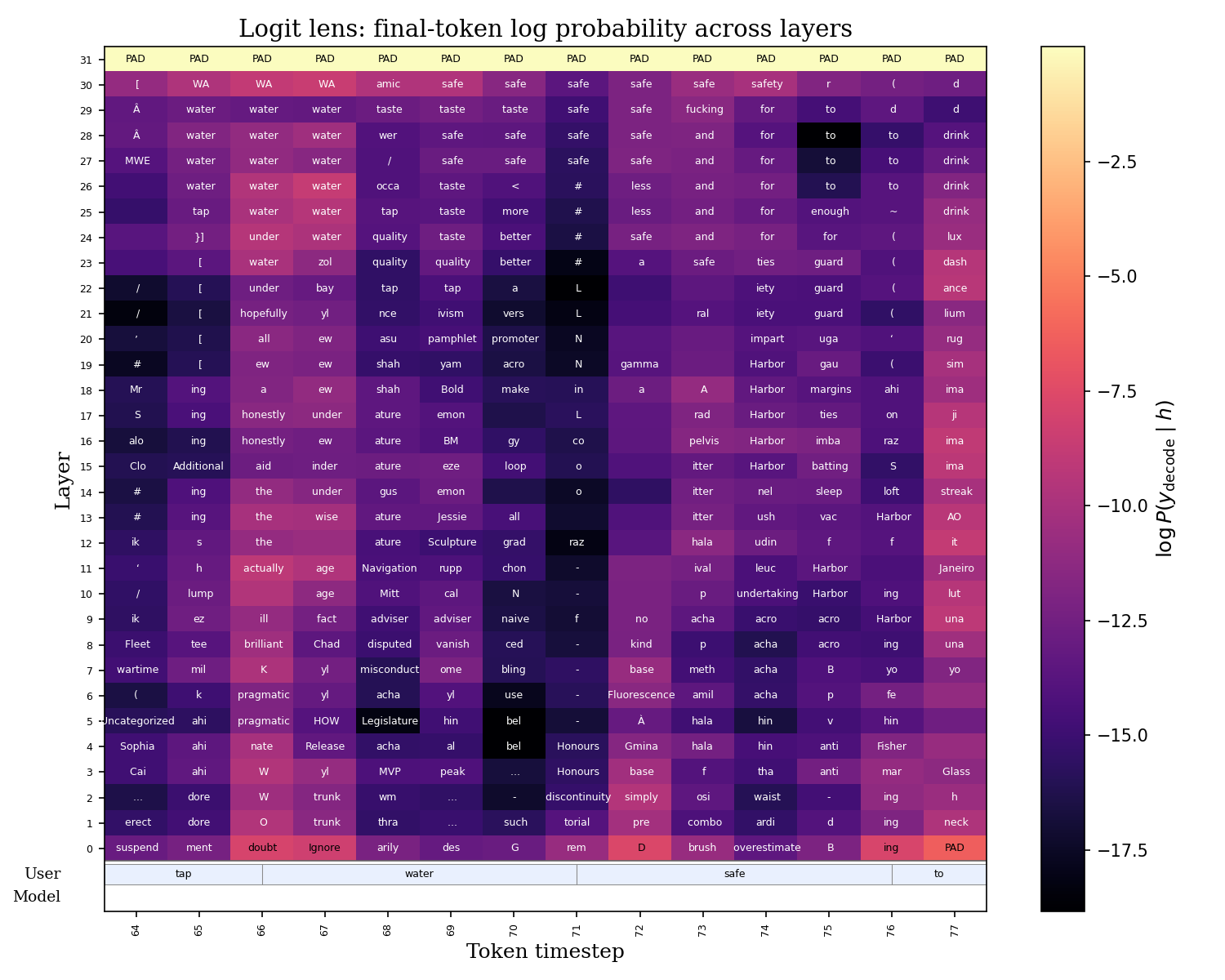}
    \caption{Additional logit-lens decoding example during a listening segment. The user input is ``How does water treatment make tap water safe to drink in modern cities?'' Intermediate layers decode tokens that anticipate or semantically track the incoming user stream: around ``tap,'' decoded tokens include ``water''; around ``water,'' decoded tokens include ``quality,'' ``safe,'' and ``tastes''; around ``safe,'' decoded tokens include ``to,'' ``for,'' and ``safety''; and around ``to,'' decoded tokens include ``drink.'' This provides further qualitative evidence that hidden states can track user-side continuations during listening.}
    \label{fig:logit_len_decode_listen_additional}
\end{figure}

Figure~\ref{fig:logit_len_decode_listen} shows that, during listening, intermediate layers often predict continuations of the incoming user utterance rather than only the model-side output token. For example, when the user-side phrase is ``their pros and cons,'' decoded tokens include ``pro,'' ``and,'' and ``cons,'' which anticipate upcoming user-side content. The decoded tokens may also be semantically related to the ongoing utterance even when they do not exactly match the next token. For example, at the timestep corresponding to the input token ``explain,'' the decoded tokens include ``why,'' ``how,'' and ``personal,'' which are relevant continuations. We highlight several representative examples in Table~\ref{tab:logit_lens_examples_appendix}. An additional layer-wise logit-lens decoding example is provided in Figure~\ref{fig:logit_len_decode_listen_additional}.

\subsection{Logit-Lens Decoding During Model Speech}

\begin{figure}[htbp]
    \centering
    \includegraphics[width=\linewidth]{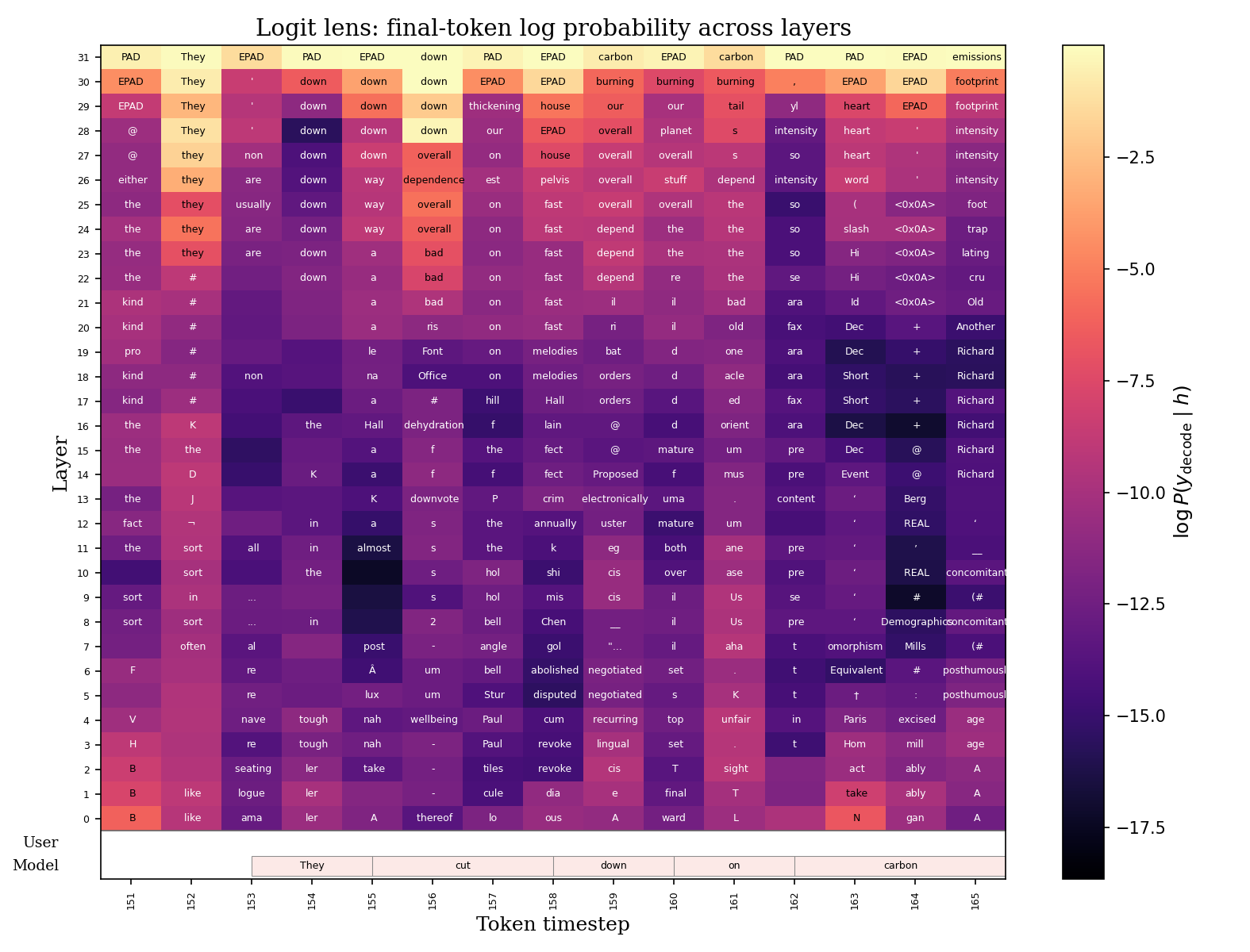}
    \caption{Logit-lens decoding of PersonaPlex hidden states during a model speaking segment. Compared with the listening segment in Figure~\ref{fig:logit_len_decode_listen}, the speaking segment shows stronger alignment with the model-side output stream across a broader range of layers, consistent with a generative state.}
    \label{fig:logit_len_decode_speak}
\end{figure}

\begin{figure}[htbp]
    \centering
    \includegraphics[width=\linewidth]{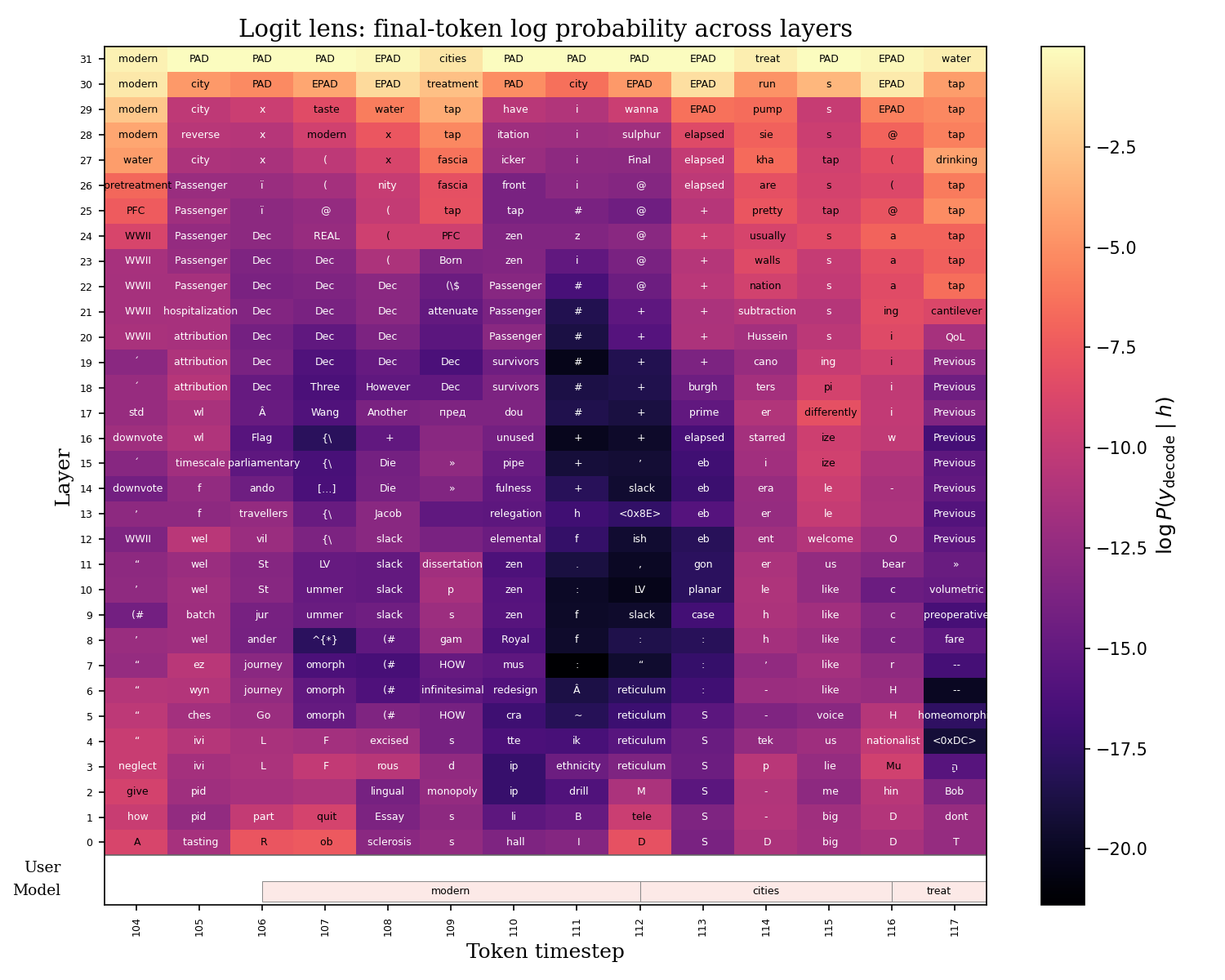}
    \caption{Additional logit-lens decoding example during a model speaking segment. This example corresponds to the model response beginning with ``Modern cities treat water...'' after the user query shown in Figure~\ref{fig:logit_len_decode_listen_additional}. The decoded tokens follow the model-side output stream, providing further qualitative evidence of generative-state alignment during speaking.}
    \label{fig:logit_len_decode_speak_additional}
\end{figure}

Figure~\ref{fig:logit_len_decode_speak} shows the complementary pattern during model speech. Intermediate hidden states assign higher projected probability to model-side text tokens, and decoded tokens more directly follow the model output stream. Some timesteps still have lower model-text probability because recent FD-SLMs often distribute text-token and audio-token generation across different frames; during audio-generation frames, the model-side text token may be \texttt{<PAD>} or \texttt{<EPAD>}. An additional layer-wise logit-lens decoding example is provided in Figure~\ref{fig:logit_len_decode_speak_additional}.

Together, Figures~\ref{fig:logit_len_decode_listen} and~\ref{fig:logit_len_decode_speak} provide qualitative evidence for stream-specific predictive focus: hidden states tend to track the incoming user stream during listening and the model-side output stream during speaking. This supports the interpretation of $S_{\mathrm{perc}}(t)$ and $S_{\mathrm{gen}}(t)$ in Section~\ref{sec:gen_per_state} as indicators of perceptive and generative states, respectively.

\section{Steering Parameter Analysis}
\label{app:steering_param_sweep}

\begin{figure}[htbp]
    \centering
    \begin{minipage}{0.48\textwidth}
        \centering
        \includegraphics[width=\linewidth]{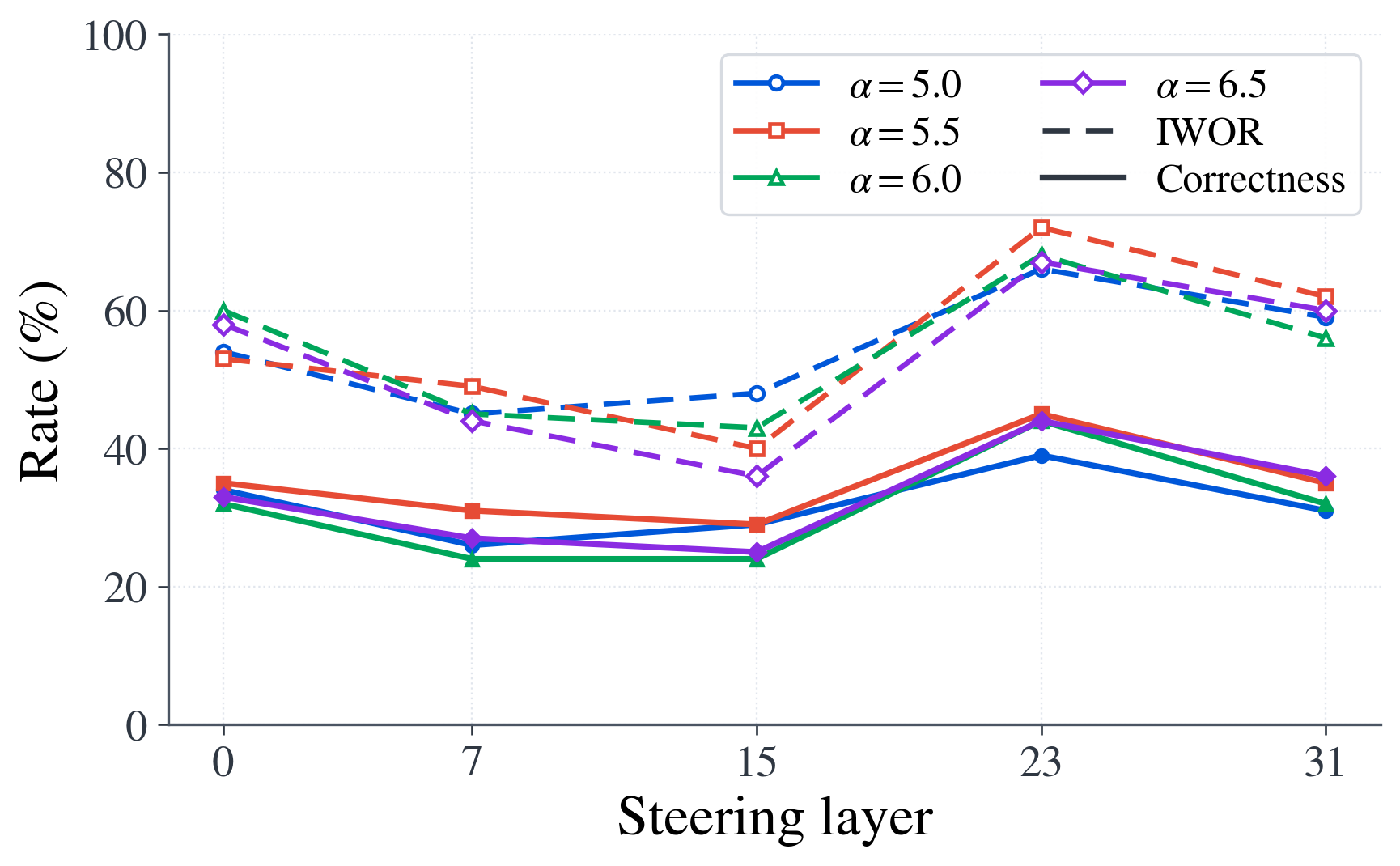}
        \caption{Correctness and IWOR across steering layers for different steering strengths $\alpha$ on PersonaPlex.}
        \label{fig:iws_rate_layers}
    \end{minipage}
    \hfill
    \begin{minipage}{0.48\textwidth}
        \centering
        \includegraphics[width=\linewidth]{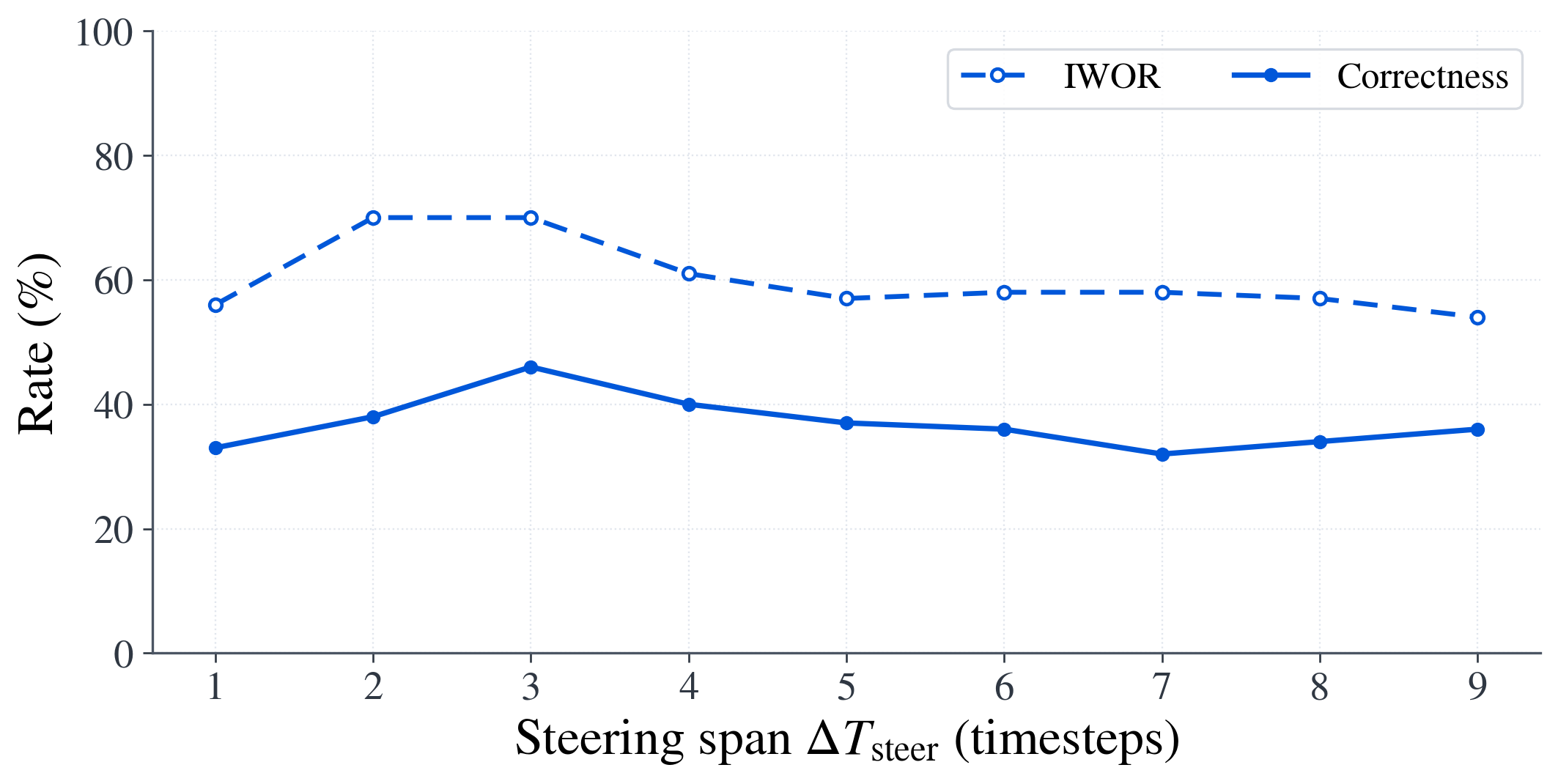}
        \caption{Correctness and IWOR across steering spans $\Delta T_{\mathrm{steer}}$ on PersonaPlex, with the steering layer fixed to 23 and $\alpha=5.5$. At $\Delta T_{\mathrm{steer}}=3$, both metrics achieve the best performance.}
        \label{fig:iws_rate_span}
    \end{minipage}
\end{figure}

\paragraph{Steering layer and strength $\alpha$.}
We investigate how the steering layer and steering strength $\alpha$ affect ZBB performance. We perform a grid search over candidate steering layers and values of $\alpha$ on PersonaPlex. As shown in Figure~\ref{fig:iws_rate_layers}, steering is most effective at layer 23 across the tested values of $\alpha$. The best configuration is achieved at $\alpha=5.5$, where correctness reaches 0.45 and IWOR reaches 0.72.

\paragraph{Steering span $\Delta T_{\mathrm{steer}}$.}
We further investigate how the steering span affects ZBB performance. For this scan, we fix the steering layer to 23 and the steering strength to $\alpha=5.5$. As shown in Figure~\ref{fig:iws_rate_span}, short steering spans already improve both correctness and IWOR over the interruption condition in Section~\ref{sec:zbb_eval_results}, while a span of 3 timesteps achieves the best overall performance. Longer spans gradually reduce performance, suggesting that steering is most effective when applied briefly at the interruption onset rather than throughout the interrupted utterance.
\section{Attention Recovery After Steering}
\label{app:attention_recovery}

Given that activation steering improves both correctness and IWOR, we further examine whether it changes attention allocation after interruption. Specifically, we measure how strongly subsequent timesteps attend back to earlier timesteps in the interrupting user input.

We compute the average attention weight assigned to the input at timestep $t$ by the subsequent $n$ timesteps at the attention layer of interest. Let $w_j(t, \tau)$ denote the attention weight from the query at timestep $\tau$ to the key at timestep $t$ in attention head $j$, and let $\mathcal{H}$ denote the set of attention heads in this layer. We define $s_t$ as the average attention score assigned to timestep $t$ over the next $n$ timesteps, averaged across all attention heads:
\begin{equation}
    s_t =
    \frac{1}{n|\mathcal{H}|}
    \sum_{\tau=t+1}^{t+n}
    \sum_{j \in \mathcal{H}}
    w_j(t, \tau).
    \label{eqn:subseq_attn_weight}
\end{equation}

This metric $s_t$ quantifies how strongly later hidden states attend back to the user input at timestep $t$. We use it to examine whether injecting the perception vector $\mu_{g \to p}$ restores attention to the beginning of the interrupting utterance.

We compute $s_t$ on ZBB examples under three conditions: \textit{no-interruption}, \textit{interruption}, and \textit{interruption with steering}. The heatmaps are aligned to the beginning of the zero-buffer query, allowing us to compare how much attention the model allocates to the earliest timesteps of the interruption.

\begin{figure}[htbp]
    \centering
    \includegraphics[width=0.9\textwidth]{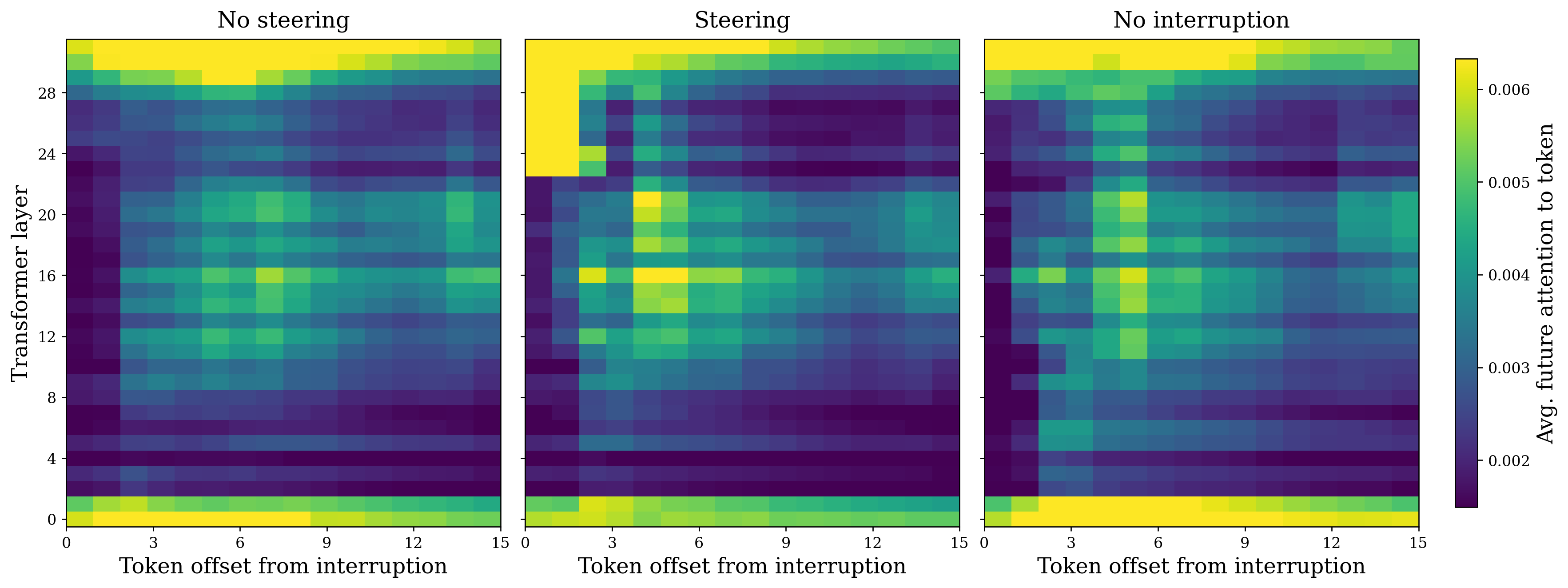}
    \caption{
    Attention recovery after steering. Heatmaps show the average attention weight assigned to each interruption timestep $t$ by subsequent timesteps at varying offsets. Attention around the 5th timestep corresponds to the first semantic word of the zero-buffer query. \textbf{Left:} In the \textit{interruption} condition, attention to the beginning of the zero-buffer query is reduced, consistent with degraded correctness and IWOR. \textbf{Middle:} In the \textit{interruption with steering} condition, injecting the perception vector $\mu_{g \to p}$ restores attention to the earliest interruption timesteps. \textbf{Right:} In the \textit{no-interruption} condition, the model allocates strong attention to the beginning of the zero-buffer query.
    }
    \label{fig:attention_subseq}
\end{figure}

Figure~\ref{fig:attention_subseq} shows that $s_t$ decreases in the \textit{interruption} condition, especially near the beginning of the zero-buffer query. After injecting the perception vector, $s_t$ in the \textit{interruption with steering} condition increases substantially relative to the \textit{interruption} condition and approaches the level of the \textit{no-interruption} condition. This result suggests that the perception vector helps restore attention to the earliest timesteps of the interrupting user input, providing additional evidence that steering mitigates state inertia at the attention level.

\section{Full-Duplex Bench Results}
\label{app:fdb_results}

We also evaluate activation steering on Full-Duplex Bench (FDB)~\cite{lin2025full_v1} to test its effect on broader full-duplex dialogue performance. We use the FDB user-interruption evaluation, which scores model responses to interruption queries on a 1--5 scale using GPT-4-Turbo. As shown in Table~\ref{tab:full_duplex_bench}, steering preserves the score within uncertainty, suggesting that the perception vector does not degrade general full-duplex response quality.

One reason is that FDB interruption queries often contain a leading filler or attention-getting phrase before the core semantic content. For example, queries such as ``Let's switch to talking about laptops'' or ``Hold on, what time is the meeting scheduled today?'' provide several initial words before the main content needed to answer the query. Therefore, unlike ZBB, FDB does not require the model to process the core semantic content immediately after interruption. By the time the core content appears, the model may have already transitioned toward the perceptive state, making FDB less sensitive to state inertia.

\begin{table}[t]
\centering
\small
\caption{Full-Duplex Bench results before and after steering, using our reproduction of the original FDB setup.}
\label{tab:full_duplex_bench}
\setlength{\tabcolsep}{6pt}
\renewcommand{\arraystretch}{1.08}
\begin{tabular}{@{}llc@{}}
\toprule
Model & Method & FDB \\
\midrule
\multirow{2}{*}{PersonaPlex}
    & Interrupt       & $3.34 \pm 0.08$ \\
    & Interrupt+Steer & $3.41 \pm 0.08$ \\
\midrule
\multirow{2}{*}{Moshi}
    & Interrupt       & $3.45 \pm 0.08$ \\
    & Interrupt+Steer & $3.36 \pm 0.08$ \\
\midrule
\multirow{2}{*}{Raon}
    & Interrupt       & $2.41 \pm 0.09$ \\
    & Interrupt+Steer & $2.41 \pm 0.09$ \\
\bottomrule
\end{tabular}
\end{table}
\section{Robustness to False Triggers}
\label{app:false_trigger}

We evaluate the robustness of activation steering to false trigger events. Since steering is applied at the detected interruption onset, an incorrect trigger could inject the perception vector when no real interruption occurs. To simulate this failure mode, we randomly inject the perception vector at incorrect timesteps while the model answers ZBB queries, and evaluate the resulting response quality using GPT-4.1-mini on a 1--5 scale.

\begin{figure}[htbp]
    \centering
    \includegraphics[width=0.7\linewidth]{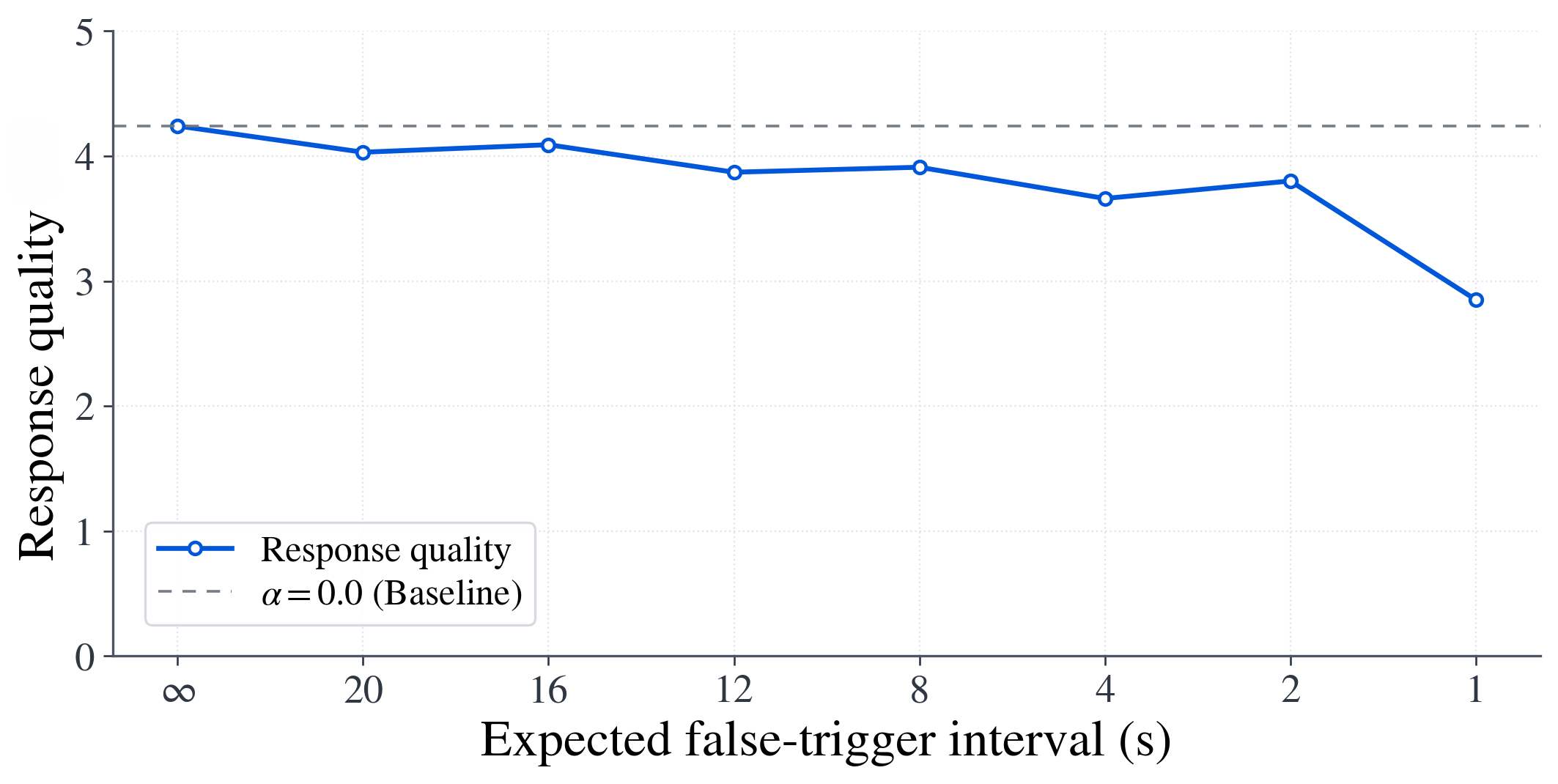}
    \caption{Response quality under false steering triggers. The x-axis represents the expected interval between false triggers. Response quality gradually decreases as false triggers become more frequent.}
    \label{fig:robustness_to_noise}
\end{figure}

As shown in Figure~\ref{fig:robustness_to_noise}, response quality degrades gradually as false triggers become more frequent. This suggests that the method is tolerant to occasional false triggers, but accurate interruption detection remains important for deployment. Semantic-aware interruption detection or VAD systems can reduce this risk by distinguishing semantically meaningful speech from non-semantic acoustic events~\cite{xia2026semantic, ball2023voice}.

\end{document}